\documentclass{article}
\usepackage{amsthm}
\usepackage{amsthm}

\PassOptionsToPackage{numbers,compress}{natbib}
\usepackage[preprint]{neurips_2020}
\usepackage{hyperref} % must be loaded before hyperxmp
\usepackage{hyperxmp}

\usepackage{amsmath, amssymb, amsthm, bm}
\usepackage{graphicx, subcaption}
\usepackage{multirow, booktabs}
\usepackage{enumitem}
\usepackage{xcolor}
\usepackage{algorithm, algpseudocode}
\usepackage{titlesec}
\usepackage{algorithm}
\usepackage{algpseudocode}
\usepackage{algorithmicx}
\usepackage{float} % If you want to use [H]
\usepackage{amsmath, amssymb}
\algrenewcommand\algorithmicindent{0.5em}

\usepackage[utf8]{inputenc}
\usepackage{amsmath}
\usepackage{amsfonts}
\usepackage{enumitem}
\usepackage{booktabs}
\definecolor{darkgreen}{rgb}{0.0, 0.5, 0.0}
\usepackage[utf8]{inputenc}
\usepackage{pgfplots}
\pgfplotsset{compat=1.18}
\usepackage{amsmath}
\usepackage{subcaption}   % after \usepackage{graphicx}
\usepackage{booktabs}   % already loaded in many templates
\usepackage{multirow}
\usepackage{graphicx}   % provides \resizebox

\titlespacing*{\section}{0pt}{0.1\baselineskip}{0.2\baselineskip}
\newcommand{\vx}{\mathbf{x}}

\newcommand{\veps}{\bm{\epsilon}}

\newcommand{\Tau}{\mathrm{T}}

\usepackage[utf8]{inputenc} % allow utf-8 input
\usepackage[T1]{fontenc}    % use 8-bit T1 fonts
\usepackage{hyperref}       % hyperlinks
\usepackage{url}            % simple URL typesetting
\usepackage{booktabs}       % professional-quality tables
\usepackage{mathtools,amssymb}
\usepackage{amsfonts}       % blackboard math symbols
\usepackage{nicefrac}       % compact symbols for 1/2, etc.
\usepackage{microtype}      % microtypography
\usepackage{pgfplots,pgfplotstable}
\pgfplotsset{compat=1.14}
\usepackage{array,colortbl}
\usepackage{xcolor}
\usepackage{algorithm,algorithmicx,algpseudocode}
\usepackage[capitalise]{cleveref}
\usepackage{caption}
\usepackage{graphbox}
\usepackage{placeins}
\usepackage{wrapfig}
\usepackage{subcaption}
\usepackage{etoolbox}

\newtoggle{hqfigures}
\togglefalse{hqfigures}  % use this for low quality figures

\title{Towards Physics-informed Diffusion for Anomaly Detection in Trajectories}

\author{%
  Arun Sharma \\
  University of Minnesota \\
  \texttt{sharm485@umn.edu} \\
  \And
  Mingzhou Yang \\
  University of Minnesota \\
  \texttt{yang7492@umn.edu} \\
  \And
  Majid Farhadloo \\
  University of Minnesota \\
  \texttt{farha043@umn.edu} \\
  \And
  Subhankar Ghosh \\
  University of Minnesota \\
  \texttt{ghosh117@umn.edu} \\
  \And
  Bharat Jayaprakash \\
  University of Minnesota \\
  \texttt{jayap015@umn.edu} \\
  \And
  Shashi Shekhar \\
  University of Minnesota \\
  \texttt{shekhar@umn.edu}
}

\begin{document}

% \begin{figure}[b!]
%   \vspace{-1em}
%   \centering
%   \includegraphics[align=c,width=0.595\textwidth]{images/celebahq256_header_image_4x4.png}\hfill
%   \includegraphics[align=c,trim=0cm 5.95cm 12cm 0cm,clip,width=0.395\textwidth]{images/cifar10_eps-fixedlarge-mse_20x20.png}
%   \caption{\small{Generated samples on CelebA-HQ $256\times 256$ (left) and unconditional CIFAR10 (right)}}
%   \label{fig:header_samples}
% \end{figure}

\maketitle

\begin{abstract}
Given trajectory data, a domain-specific study area, and a user-defined threshold, we aim to find anomalous trajectories indicative of possible GPS spoofing (e.g., fake trajectory). The problem is societally important to curb illegal activities in international waters, such as unauthorized fishing and illicit oil transfers. The problem is challenging due to advances in AI generated in deep fakes generation (e.g., additive noise, fake trajectories) and lack of adequate amount of labeled samples for ground-truth verification. Recent literature shows promising results for anomalous trajectory detection using generative models despite data sparsity. However, they do not consider fine-scale spatiotemporal dependencies and prior physical knowledge, resulting in higher false-positive rates. To address these limitations, we propose a physics-informed diffusion model that integrates kinematic constraints to identify trajectories that do not adhere to physical laws. Experimental results on real-world datasets in the maritime and urban domains show that the proposed framework results in higher prediction accuracy and lower estimation error rate for anomaly detection and trajectory generation methods, respectively. Our implementation is available at \url{https://github.com/arunshar/Physics-Informed-Diffusion-Probabilistic-Model}.

\end{abstract}

\section{Introduction}

\begin{figure}[ht!]
    \centering
    \begin{subfigure}[b]{0.48\columnwidth}
        \centering
        \includegraphics[width=\linewidth]{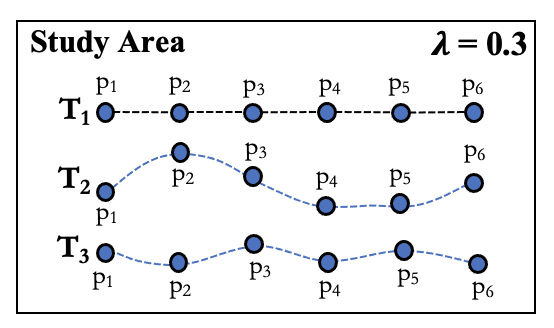} % Replace with your actual path
        \caption{Trajectory Dataset}
        \label{fig:input}
    \end{subfigure}
    \hfill
    \begin{subfigure}[b]{0.48\columnwidth}
        \centering
        \includegraphics[width=\linewidth]{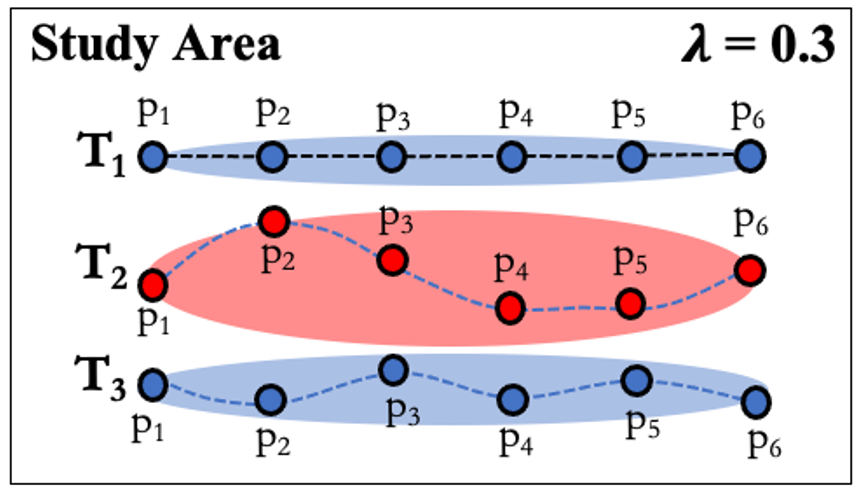} % Replace with your actual path
        \caption{Abnormal Trajectory}
        \label{fig:output}
    \end{subfigure}

    \caption{An illustration example of problem statement where trajectories $T_{1}$, $T_{2}$, $T_{3}$ are input (left) and $T_{2}$ is classified as abnormal with reconstruction threshold $\lambda \geq 0.3$}
    \label{fig:Problem Statement}
\end{figure}

Given trajectory data, a domain-specific study area (e.g., road network, maritime waters), and a user-defined reconstruction threshold, our aim is to identify anomalous trajectories indicative of possible deception-based activity or behavior. Deception-based anomalous behavior occurs when an object intentionally broadcasts spurious or fake locations to conceal its movements for potential illicit activities and deceive the end-user. These false signals do not conform to the historical mobility patterns of nearby entities and are not physically feasible resulting in higher reconstruction errors by generative models. For instance, Figure \ref{fig:input} shows trajectories $T_{1}$,$T_{2}$, and $T_{3}$ where trajectory $T_{2}$ exhibits significantly different mobility behavior compared to trajectories $T_{1}$ and $T_{3}$ in terms of angle and physical parameters (e.g., speed) etc. In Figure \ref{fig:output},  $T_{2}$’s behavior is flagged as abnormal due to its deviation from historical trajectory data and physically feasible attributes (e.g., bearing, speed), as reflected by its reconstruction error $\lambda$, which exceeds the threshold \textbf{0.3}. In this paper, we investigate a physics-informed diffusion probabilistic model (Pi-DPM) that effectively analyzes spatiotemporal characteristics along with historical trajectories and leverage physical kinematic models to detect anomalous behavior that does not abide by the laws of physics.

\begin{figure}[htb]
    \centering
    \includegraphics[width=0.95\textwidth]{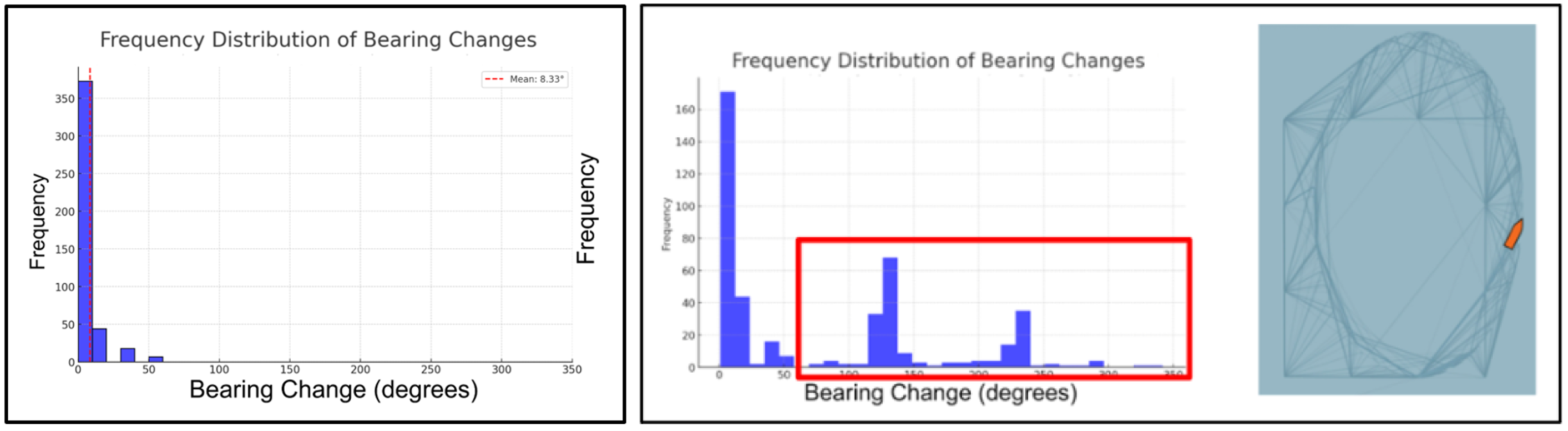}
    \captionsetup{justification=centering}
    \caption{A case study of an oil tanker Cathay Phoenix whose bearing distribution (right) is significantly different towards normal behavior (left).}
    \label{fig:normal}
    \label{fig:motivation}
\end{figure}

GPS Spoofing has significant implications for homeland security, public health and safety, enforcement of UN sanctions, maritime regulations, etc. An example of deception-based activity that put the public at risk occurred in 2020 when a Berlin artist used 99 phones to manipulate Google Maps to reroute traffic from a residential street \cite{hern2020berlin}. In another case, The New York Times \cite{triebert2023fake} reported on an oil tanker, the Cathay Phoenix, that broadcast fake location signals near the Japan Sea, indicating it was making frequent abrupt changes in its direction. While the typical bearing change distribution for tankers generally follows a smooth path (Figure \ref{fig:motivation} on left), the Cathay Phoenix exhibited sudden directional shifts within a confined geographic region, as indicated by its bearing change distribution (red box in Figure \ref{fig:motivation} in the right). The ship’s actual location at the time was a Russian port, where it was loading oil for shipment to China to evade US sanctions. 

Identifying anomalous trajectories arising from possible GPS spoofing is challenging due to advances in AI generated in deep fakes generation (e.g., additive noise, fake trajectories). Moreover, the scarcity of ground truth data and known attack models hinders the development of accurate and physically plausible anomaly detection methods. Finally, machine learning models provide statistical estimation and often fail to integrate real physical constraints when generating trajectories. This limits their interpretability and explainability, making it hard for human analysts to manually inspect such fake trajectories.

\begin{figure}[htb]
    \centering
    \includegraphics[width=0.55\textwidth]{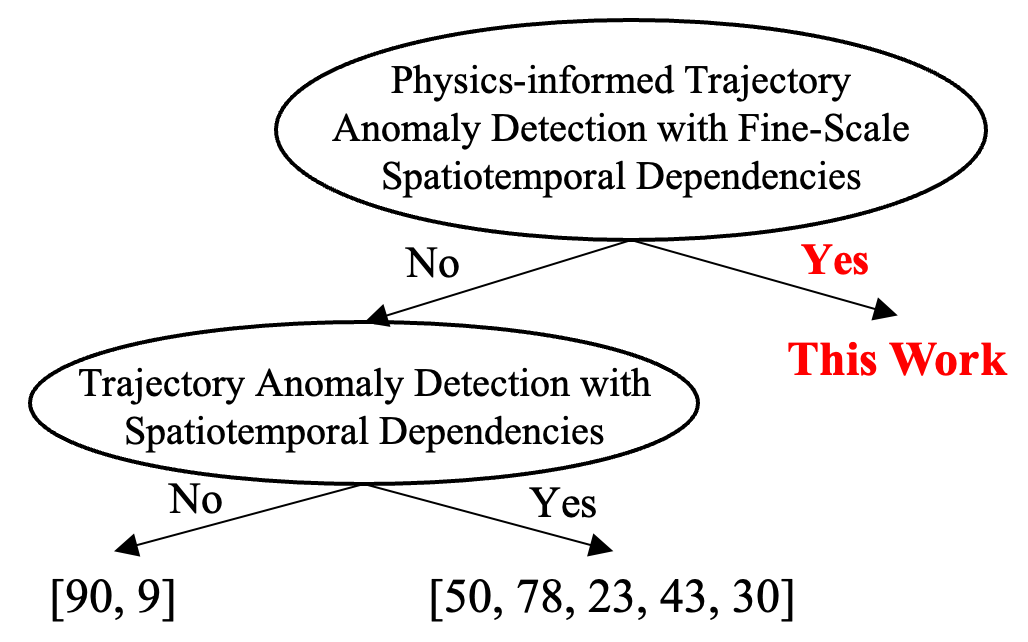}
    \captionsetup{justification=centering}
    \caption{Limitations of Related Work}
    \label{fig:normal}
    \label{fig:related_work}
\end{figure}

Recent literature on deep learning-based anomalous detection methods for trajectories, uses generative models to learn spatial and temporal feature representations for classifying anomalous trajectories. However, these methods are limited in capturing anomalous trajectories at a coarser level (e.g., aggregated by asynchronous routes or trips) \cite{zhang2011ibat,chen2013iboat}, which restricts their ability to learn real-time spatiotemporal dependencies at a finer granularity for investigating complex behavioral mobility patterns. Additionally, these models \cite{liu2020online,wang2024multi,gao2023open,li2024causaltad,han2022deeptea} do not account for physics-based parameters (e.g., rate of turn, acceleration), making it challenging to investigate real-time vehicle dynamics, which are crucial for detecting rare but significant anomalous behaviors such as GPS spoofing. In this work, we propose a Diffusion-based Probabilistic Model that incorporates real-time spatiotemporal dependencies and physics-based parameters \cite{polack2017kinematic} to enforce kinematic constraints at a finer granularity. This ensures that detected anomalies reflect physically infeasible behaviors rather than purely statistical outliers. Figure \ref{fig:related_work} shows the overall comparison with related work.\\

\vspace*{-0.8em} % tighten space after the float

\textbf{Our contributions are as follows:} \vspace{-0.5em}
\begin{itemize}
    \item We propose a novel Physics-informed Diffusion Probabilistic Model (Pi-DPM) that adheres fine-scale spatiotemporal dependencies and kinematic constraints for generating physically plausible trajectories.
    \item We introduce Physics-informed Regularization to more accurately capture reconstruction loss in the sampling phase.   
    \item We compare the Pi-DPM against state-of-the-art trajectory generation and anomaly detection methods and conduct comprehensive evaluations on different datasets.
\end{itemize}

\textit{\textbf{Scope:}} This paper focuses on a simple kinematic model with 3 degrees-of-freedom. The model excludes complex physics due to dataset limitations. The model is applied to maritime and road networks, using Gaussian noise \cite{ho2020} for the denoising process.

\textit{\textbf{Organization:}} The rest of the paper is organized as follows. Section \ref{sec:preliminary} introduces key concepts and formally defines the problem. \ref{sec:Proposed Approach} describes the proposed model in detail. In Section \ref{sec:Experiment}, we evaluate the proposed model experimentally. A review of related work is given in Section \ref{related_work} followed by a brief discussion in Section \ref{sec:Discussion}. Finally, Section \ref{sec:Conclusion} concludes the paper and outlines future work.

\section{Problem Formulation}\label{sec:preliminary}

\subsection{Notations and Definitions}
A \textbf{spatial trajectory} is a collection of location traces generated by a moving object, typically represented as a chronologically ordered sequence of points, $p_1 \to p_2 \to \dots \to p_n$, where each point $p$ consists of $(lat, long, t, v)$. For instance, in Figure \ref{fig:input}, the trajectories $T_1$, $T_2$, and $T_3$ consist of ordered location traces, where each trajectory $T_i$ is defined as $[p_1, \dots, p_n]$. A collection of such trajectories forms a trajectory dataset: $T = [T_1, T_2, \dots, T_n]$. 

\begin{figure}[htb]
    \centering
    \includegraphics[width=0.8\textwidth]{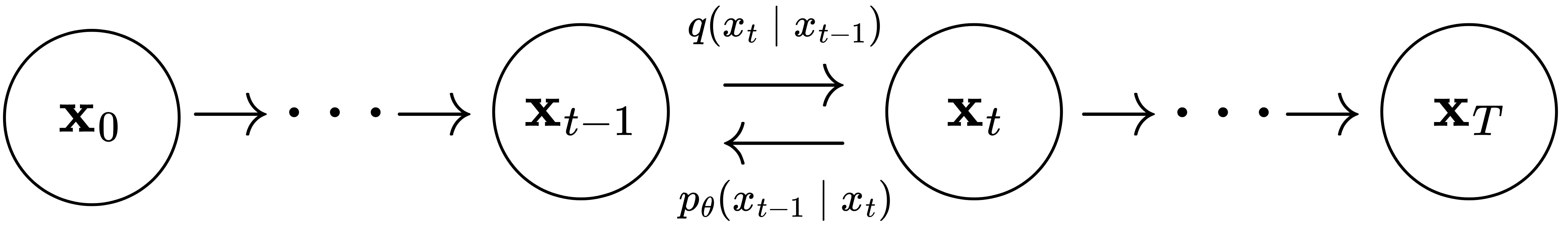}
    \captionsetup{justification=centering}
    \caption{Forward and reverse diffusion processes}
    \label{fig:DiffusionToy}
\end{figure}

A \textbf{forward process} progressively introduces ($q$) Gaussian noise to $\vx_0$ over $T$ iterations \cite{ho2020} as described by:

\begin{equation}
    q(\mathbf{x}_{1:T} \mid \mathbf{x}_0) = \prod_{t=1}^{T} q(\mathbf{x}_{t} \mid \mathbf{x}_{t-1})~,
\end{equation}
\begin{equation}
    q(\mathbf{x}_{t} \mid \mathbf{x}_{t-1}) = \mathcal{N}(\mathbf{x}_{t} \mid \sqrt{\alpha_t}\, \mathbf{x}_{t-1}, (1 - \alpha_t) \mathbf{I} )~,
\end{equation}

Here, the hyperparameters $\alpha_{1:T}$ are constrained between $0$ and $1$, representing the noise variance introduced at each iteration. The variable $\vx_{t-1}$ is scaled down by a factor of $\sqrt{\alpha_t}$ to keep the variance of the random variables finite. Additionally, deriving $\vx_t$ from $\vx_0$ can be streamlined using Equation \ref{eq:diffusion-marginalized}.:

\begin{equation}
    q(\vx_t \mid \vx_0) ~=~ \mathcal{N}(\vx_t \mid \sqrt{\bar{\alpha}_t}\, \vx_0, (1-\bar{\alpha}_t) \bm{I})~,
\label{eq:diffusion-marginalized}
\end{equation}
where $\bar{\alpha}_t = \prod_{i=1}^t \alpha_i$. In addition,   one can derive the posterior distribution of $\vx_{t-1}$ given $(\vx_0, \vx_t)$ as
\begin{equation}
\begin{aligned}
    q&(\vx_{t-1} \mid \vx_0, \vx_t) = \mathcal{N}(\vx_{t-1} \mid \bm{\mu}, \sigma^2 \bm{I})\\
    &\bm{\mu} =  \frac{\sqrt{\bar\alpha_{t-1}}\,(1-\alpha_t)}{1-\bar{\alpha}_{t}}\, \vx_0 \!+\! \frac{\sqrt{\alpha_t}\,(1-\bar{\alpha}_{t-1})}{1-\bar\alpha_t}\vx_t\\
    &\sigma^2 = \frac{(1-\bar\alpha_{t-1})(1-\alpha_t)}{1-\bar{\alpha}_{t}}~.
\end{aligned}
\label{eq:posteriror-ytm1}
\end{equation}

Figure \ref{fig:DiffusionToy} presents a detailed view of a diffusion model, where in the forward process, isotropic Gaussian noise is progressively added to a signal through a predetermined Markov chain, denoted by $q(\vx_t \mid \vx_{t-1})$, generating a series of intermediate noise variances referred as \textit{forward} diffusion.

\begin{figure}[H]
  \centering
  %--- Left: Training --------------------------------------------------------
  \begin{minipage}[t]{0.48\linewidth}
    \begin{algorithm}[H]
      \caption{Training}\label{alg:training}
      \small
      \begin{algorithmic}[1]
        \Repeat
          \State $(\mathbf{x}_0) \sim p(\mathbf{x})$
          \State $\alpha \sim p(\alpha)$
          \State $\bm{\epsilon} \sim \mathcal{N}(\mathbf{0}, \mathbf{I})$
          \State Take a gradient‐descent step on
          \Statex\hspace{1em}$\nabla_\theta\!\left\lVert
            \bm{\epsilon}
            - f_\theta\!\bigl(\mathbf{y},
              \sqrt{\bar{\alpha}}\,\mathbf{x}_0
              + \sqrt{1-\bar{\alpha}}\,\bm{\epsilon},
              \bar{\alpha}\bigr)
            \right\rVert_p^p$
        \Until{converged}
      \end{algorithmic}
    \end{algorithm}
  \end{minipage}
  \hfill
  %--- Right: Inference ------------------------------------------------------
  \begin{minipage}[t]{0.48\linewidth}
    \begin{algorithm}[H]
      \caption{Inference}\label{alg:inference}
      \small
      \begin{algorithmic}[1]
        \State $\mathbf{x}_T \sim \mathcal{N}(\mathbf{0}, \mathbf{I})$
        \For{$t = T, T{-}1, \dotsc, 1$}
          \State $\mathbf{z} \sim \mathcal{N}(\mathbf{0}, \mathbf{I})$ \textbf{if} $t > 1$ \textbf{else} $\mathbf{z} = \mathbf{0}$
          \State $\displaystyle
            \mathbf{x}_{t-1} =
            \frac{1}{\sqrt{\alpha_t}}
              \bigl(
                \mathbf{x}_t -
                \tfrac{1-\alpha_t}{\sqrt{1-\bar{\alpha}_t}}\,
                f_\theta(\mathbf{x}_t,\bar{\alpha}_t)
              \bigr) +
            \sqrt{1-\alpha_t}\,\mathbf{z}$
        \EndFor
        \State \textbf{return} $\mathbf{x}_0$
      \end{algorithmic}
    \end{algorithm}
  \end{minipage}
  \caption{Side‐by‐side training (\textbf{left}) and inference (\textbf{right}) procedures for the diffusion model.}
\end{figure}

A \textbf{reverse process ($p_\theta$)} is the \textit{reverse} of a forward diffusion process in that it begins with a completely noisy state, $\vx_T \sim \mathcal{N}(\bm{0}, \bm{I})$. The model then incrementally refines the trajectories through successive iterations $(\vx_{T-1}, \vx_{T-2}, \dots, \vx_0)$. In contrast to the forward process $q$, $p_\theta$ goes in the reverse direction starting from Gaussian noise $\bm{x}_{T}$:
\begin{eqnarray}
    p_\theta(\vx_{0:T}) &=& p(\vx_T) \prod\nolimits_{t=1}^T p_\theta(\vx_{t-1} | \vx_t), \\
    p(\vx_T) &=& \mathcal{N}(\vx_T \mid \bm{0}, \bm{I}), \label{eq:mutheta}\\
    p_\theta(\vx_{t-1} | \vx_{t}) &=& \mathcal{N}(\vx_{t-1} \mid \mu_{\theta}({\vx}_{t}, \alpha_t), \sigma_t^2\bm{I})~, \label{eq:reverse_process}
\end{eqnarray}

where \( p_\theta(\mathbf{x}_{0:T}) \) is the joint probability distribution of \( (\mathbf{x}_0, \dots, \mathbf{x}_T) \) and \( p(\mathbf{x}_T) \) is a Gaussian distribution. The mean \( \mu_\theta(\mathbf{x}_t, t) \) accepts \( \mathbf{x}_t \) and \( t \)  as input and is obtained by learning while the variance term $\sigma_t^2\bm{I}$ follows the inference process defined using isotropic Gaussian distributions, $p_\theta(\vx_{t-1} | \vx_{t})$ which are learned. If the noise variance in the forward process steps is minimized, \emph{i.e.}, $\alpha_{1:T} \approx 1$, the resulting optimal reverse process in \eqref{eq:reverse_process} $p_\theta(\vx_{t-1} | \vx_{t})$ will closely approximate a Gaussian distribution~\cite{sohl2015deep}. Moreover, it is necessary for $1 - \alpha_T$ to be large enough to ensure that the distribution of $\mathbf{x}_T$ closely aligns with the prior distribution $p(\mathbf{x}_T) = \mathcal{N}(\mathbf{x}_T | \mathbf{0}, \mathbf{I})$, which is a standard Gaussian distribution with mean zero and identity covariance matrix. Here we designed $f_\theta$ to predict $\veps$, including $x_t$. Consequently, we can estimate $\vx_0$ by rearranging the terms as shown in \eqref{eq:noisy-y}:

\begin{equation}
\label{eq:noisy-y}
    \hat{\vx}_0 = \frac{1}{\sqrt{\bar{\alpha_t}}} \left( \vx_t - \sqrt{1 - \bar{\alpha_t}}\, f_{\theta}( \vx_{t}, \bar{\alpha_t}) \right)~.
\end{equation}

Finally, we insert $\hat{\vx}_0$ into the posterior distribution of $q (\vx_{t-1} | \vx_0, \vx_t)$, which parameterizes the mean of $p_\theta(\vx_{t-1} | \vx_t)$ in Equation \ref{eq:12}. We set the variance of $p_\theta(\vx_{t-1}|\vx_t)$ to $(1 - \alpha_t)$, which is the default variance as determined by the variance of the forward process \cite{ho2020denoising}:
\begin{equation}
\label{eq:12}
    \mu_{\theta}({\vx}_{t}, \bar{\alpha_t}) = \frac{1}{\sqrt{\alpha_t}} \left( \vx_t - \frac{1-\alpha_t}{ \sqrt{1 - \bar{\alpha_t}}} f_{\theta}(\vx_{t}, \bar{\alpha_t}) \right)~,
\end{equation}

Additionally, we carry out iterative refinement at each iteration as described in Equation \ref{eq:13} (where $\veps_t \sim \mathcal{N}(0,1)$):
\begin{equation}
\label{eq:13}
\vx_{t-1} \leftarrow \frac{1}{\sqrt{\alpha_t}} \left( \vx_t - \frac{1-\alpha_t}{ \sqrt{1 - \bar{\alpha_t}}} f_{\theta}(\vx_{t}, \bar{\alpha_t}) \right) + \sqrt{1 - \alpha_t}\veps_t~,
\end{equation}

An \textbf{anomalous vehicle trajectory} in a traffic scene deviates from typical patterns in aspects such as position, orientation, and speed. These deviations often arise from violations of traffic rules or physical driving constraints, such as sudden or extreme changes in movement due to complex or unpredictable conditions. Common examples include drifting off, prolonged stops, or driving in the opposite direction. 

A \textbf{reconstruction threshold} $\lambda$ is the error margin derived from comparing the divergence between reconstructed and original input trajectories using a generative model. A trajectory is classified as an \textbf{anomalous trajectory} $T_a$ if its reconstruction error exceeds or equals $\lambda$, indicating significant deviations in latitude, longitude, speed, or heading compared to neighborhood trajectories. For instance, in Figure \ref{fig:output}, trajectory $T_2$ in red exhibits substantial deviations relative to $T_1$ and $T_3$, with a reconstruction error $\geq 0.3$, and is therefore flagged as anomalous.

Table \ref{tab:notation_table} enumerates all mathematical symbols and abbreviations adopted in the manuscript. By collating trajectory variables, kinematic-model parameters, diffusion hyper-parameters, and loss terms in one place, the table establishes a uniform notation that is adhered to throughout the paper.

\begin{table}[htb]
\caption{Summary of Mathematical Notations and Symbols}
\label{tab:notation_table}
\footnotesize
\setlength{\tabcolsep}{3pt}
\centering  
\begin{tabular}{|l|l|}
\hline
\textbf{Notation} & \textbf{Description}\\
\hline
$p_t$ & Location sample at time $t$: $(\text{lat},\text{lon},v,\psi)$ \\
\hline
$\vx_0$ & Clean (noise–free) trajectory fed into the forward diffusion \\
\hline
$\vx_t,\;\vx_{t-1}$ & Noisy trajectory after $t$ / $t{-}1$ diffusion steps \\
\hline
$\vx_T$ & Pure‐noise trajectory at the end of the forward process \\
\hline
$\hat{\vx}_0$ & Trajectory reconstructed by Pi-DPM (output of reverse process) \\
\hline
$v,\;\hat v$ & True / reconstructed speed (m s$^{-1}$) \\
\hline
$a,\;\hat a$ & True / reconstructed longitudinal acceleration (m s$^{-2}$) \\
\hline
$\psi,\;\hat\psi$ & Heading (yaw) angle / reconstructed heading (rad) \\
\hline
$\delta$ & Steering angle in the kinematic bicycle model (rad) \\
\hline
$L$ & Wheel-base of the vehicle (m) \\
\hline
$\kappa,\;\hat\kappa$ & Curvature, $\kappa=\tan\delta / L$ / reconstructed curvature \\
\hline
$R$ & Turning radius, $R=L/\tan\delta$ \\
\hline
$T$ & Total number of diffusion time-steps ($T{=}1000$ in experiments) \\
\hline
$t$ & Current diffusion step, $1\le t\le T$ \\
\hline
$\alpha_t$ & Forward noise–scaling coefficient at step $t$ \\
\hline
$\bar\alpha_t$ & Cumulative product $\prod_{i=1}^{t}\alpha_i$ \\
\hline
$\beta_t$ & Forward variance increment, $\beta_t = 1-\alpha_t$ \\
\hline
$\sigma_t$ & Fixed variance in reverse step $p_\theta(\vx_{t-1}\!\mid\!\vx_t)$ \\
\hline
$\bm{\epsilon},\;\epsilon_t$ & Standard Gaussian noise vector / sample at step $t$ \\
\hline
$\mathbf{z}$ & Auxiliary noise drawn during sampling when $t>1$ \\
\hline
$f_\theta$ & Neural network predicting $\bm{\epsilon}$ or $\mu_\theta$ in reverse process \\
\hline
$\lambda$ & Reconstruction-error threshold for anomaly detection \\
\hline
$w_{1\text{–}4}$ & Weights for KBM constraints in physics loss $\mathcal{L}_{\text{Phy}}$ \\
\hline
$\gamma_{1\text{–}3}$ & Weights for $\mathcal{L}_{\text{VLB}}$, $\mathcal{L}_{\text{Rec}}$, $\mathcal{L}_{\text{Phy}}$ \\
\hline
$\mathcal{L}_{\text{VLB}}$ & Variational lower-bound (diffusion) loss \\
\hline
$\mathcal{L}_{\text{Rec}}$ & Reconstruction loss \\
\hline
$\mathcal{L}_{\text{Phy}}$ & Physics-based regularization loss (KBM constraints) \\
\hline
$\mathcal{L}_{\text{Pi-DPM}}$ & Total loss $\gamma_1\mathcal{L}_{\text{VLB}}+\gamma_2\mathcal{L}_{\text{Rec}}+\gamma_3\mathcal{L}_{\text{Phy}}$ \\
\hline
\end{tabular}
\end{table}

\begin{figure*}[t]
    \centering
    \includegraphics[width=\linewidth]{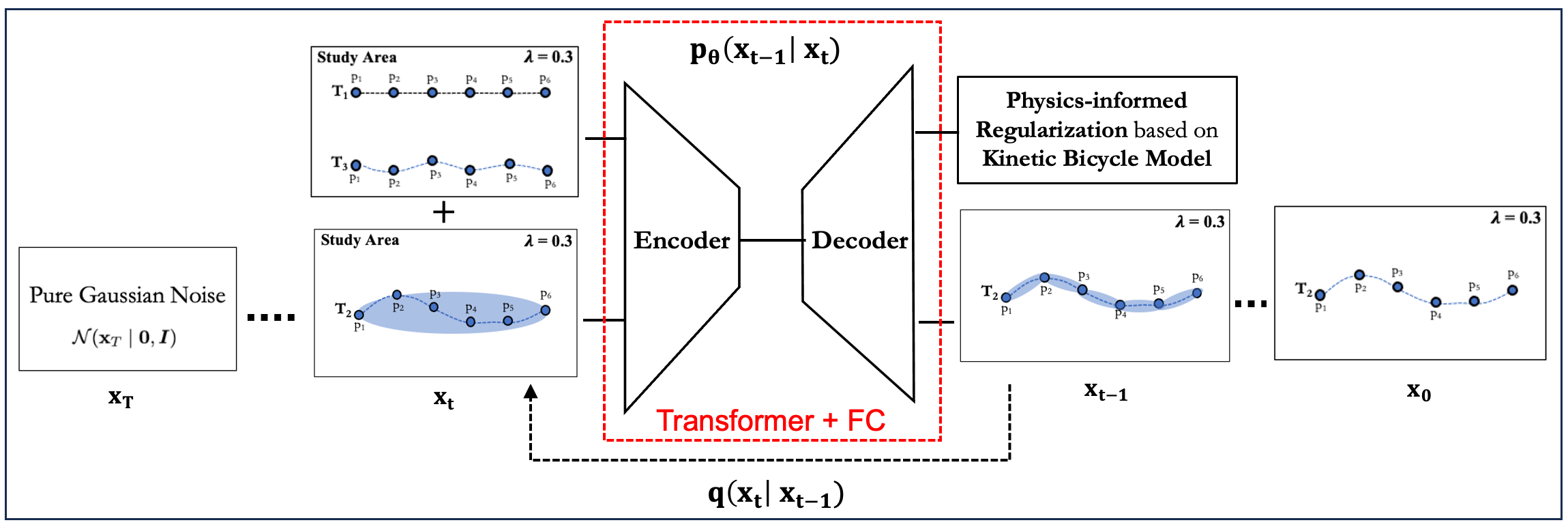}
    \caption{Physics-informed Diffusion Model (Pi-DPM) Architecture}
    \label{fig:Diffusion}
\end{figure*}

\subsection{Problem Statement}
\noindent\textbf{Input:}
\begin{itemize}[noitemsep,topsep=0pt]
    \item [--] A trajectory dataset $T$ $\in$ {$T_{1}, T_{2}, ... , T_{n}$}
    \item [--] A diffusion model
    \item [--] A reconstruction threshold, $\lambda$
\end{itemize}

\noindent\textbf{Output:} 
\begin{itemize}[noitemsep,topsep=0pt]
    \item [--] An anomalous trajectory $T_{a}$ $\geq$ $\lambda$
\end{itemize}

\noindent\textbf{Objective:} Solution quality (e.g., accuracy, F1-score)\\
\noindent\textbf{Constraints:} Data availability, ground truth

\begin{figure}[h]
    \centering
    \includegraphics[width=\linewidth]{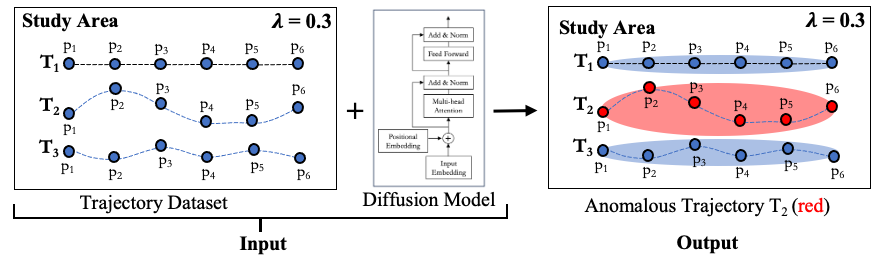}
    \caption{Problem Statement}
    \label{fig:Proposed}
\end{figure}

Figure \ref{fig:Proposed} shows three trajectories $T_{1}$, $T_{2}$, and $T_{3}$, where trajectory $T_{2}$ exhibits significantly different mobility behavior compared to $T_{1}$ and $T_{3}$ in terms of angular deviation and physical parameters such as speed resulting in a higher reconstruction error ($\lambda$) in red, as indicated in $T_{2}$'s abnormal movement behavior with $\lambda \geq 0.3$.

\section{Physics-informed Diffusion Probabilistic Model (Pi-DPM) }
\label{sec:Proposed Approach}
This section introduces our proposed model which leverages encoder-decoder-based architecture where spatial-awareness and kinematic constraints are added as a conditional input to the decoder to generate physically plausible trajectories. 
\vspace{-0.5em}
\subsection{Proposed Architecture}
Our encoder–decoder architecture is designed to reconstruct physically plausible trajectories while distinguishing anomalies indicative of deceptive behaviors. Departing from conventional symmetric autoencoders, Figure \ref{fig:Transformer} shows a framework which employs an asymmetric structure by a \textbf{context-informed encoder} extracts spatio-temporal features using a transformer-based attention mechanism, while a \textbf{physics-informed decoder} reconstructs trajectories under explicit kinematic constraints. This separation enables the model to retain contextual richness during encoding and enforce physical realism during the denoising phase.

\textbf{Training and Inference:}  
Our proposed model extends the denoising diffusion probabilistic model (DDPM)~\cite{ho2020} by conditioning the reverse process on the local spatio-temporal context derived from neighboring trajectories. In multi-agent systems such as vehicles, preserving inter-agent dependencies is critical for accurate modeling and detection of abnormal behaviors. To this end, we introduce a transformer-based architecture which consists of two main components, a training module for learning the diffusion process, and a sampling module for trajectory reconstruction and detection. The model begins by encoding vehicle positions using two separate fully connected layers, generating embeddings that serve as inputs to the network. During training phase, we first decouple temporal and spatial interactions: \textit{temporal dependencies} are modeled using a standard transformer applied independently to each vehicle, while spatial interactions are captured through a specialized contextual encoder. For spatial encoding, we construct a topological graph where each vehicle is treated as a node, and spatial relationships are represented as edges derived from a message-passing mechanism. The spatial encoder integrates a spatial attention sublayer and a temporal convolutional sublayer to capture cross-agent dynamics over time. The outputs of both encoders are combined into latent state embeddings that encode the spatio-temporal structure of trajectories. These embeddings are used to estimate the reverse diffusion process by minimizing the variational lower bound. 

\textbf{Execution Trace:} The \textit{forward} process gradually corrupts the reconstruction error of the input trajectory by adding Gaussian noise until it converges to complete spatial randomness, denoted as $\bm{x}_{T}$. In the \textit{reverse} process, the model denoises $\bm{x}_{T}$ step-by-step to approximate the original trajectory $\Tilde{\bm{x}}_0$. At each denoising step, physics-informed decoder parameters are learned to enforce kinematic feasibility, while conditioning on surrounding trajectories to enhance spatio-temporal contextuality. Figure~\ref{fig:Diffusion} shows an example of a typical transition between states $\bm{x}_{t}$ and $\bm{x}_{t-1}$. The forward transition follows $q(\bm{x}_{t} | \bm{x}_{t-1})$, while the reverse denoising step is modeled by $p_{\theta}(\bm{x}_{t-1} | \bm{x}_{t})$. At each timestep $t$, we concatenate the current noisy trajectory $\bm{x}_{t}$ with its spatially proximate neighbors and feed them into an encoder-decoder architecture. The encoder leverages spatio-temporal attention, and the decoder outputs a denoised trajectory $\bm{x}_{t-1}$. Iterating this reverse process ultimately yields a reconstructed trajectory. For training, we adopt a piecewise noise scheduling strategy~\cite{ho2016generative}, defined as $p(\bar\alpha) = \sum_{t=1}^{T} \frac{1}{T} U(\bar\alpha_{t-1}, \bar\alpha_t)$, where a timestep $t \sim \{0, \ldots, T\}$ is uniformly sampled, followed by $\bar\alpha \sim U(\bar\alpha_{t-1}, \bar\alpha_t)$ with $T = 1000$.

\subsection{Context-Informed Encoder}  
\begin{figure}[ht!]
    \centering
    \begin{subfigure}[b]{0.35\columnwidth}
        \centering
        \includegraphics[width=\linewidth]{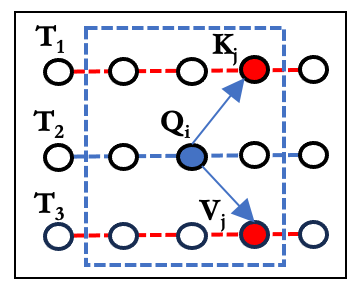}
        \caption{Spatial Attention}
        \label{fig:spatialattention}
    \end{subfigure}
    \hfill
    \begin{subfigure}[b]{0.35\columnwidth}
        \centering
        \includegraphics[width=\linewidth]{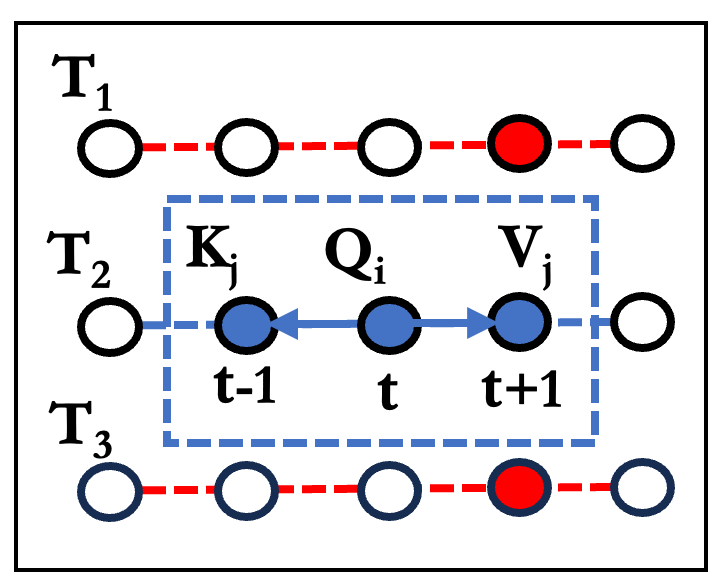}
        \caption{Temporal Attention}
        \label{fig:temporalattention}
    \end{subfigure}
    \caption{Context-informed Encoder}
    \label{fig:Encoder}
\end{figure}

\begin{figure}[h]
    \centering
    \includegraphics[width=\linewidth]{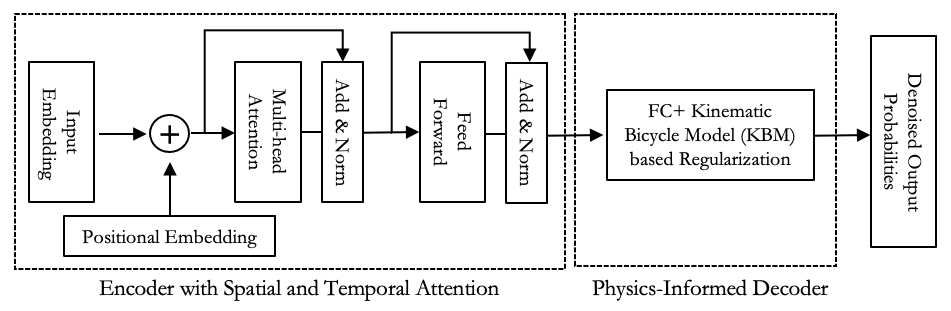}
    \caption{Proposed Encoder-Decoder Model}
    \label{fig:Transformer}
\end{figure}

To model contextual dependencies, Pi-DPM embeds object positions and applies spatial-temporal attention \cite{vaswani2017attention}. Figure \ref{fig:Encoder} shows how Contextual-Informed Encoder in Figure \ref{fig:Transformer} captures spatial and temporal dependencies via leveraging neighborhood information.

\textbf{Spatial attention} uses a sliding window at time \( t \), where the target object's trajectory features form the query \( Q \), and its neighbors contribute keys \( K \) and values \( V \). Attention weights, which are computed via softmax-scaled dot products of \( Q \) and \( K \),highlight the relevant neighbors. In Figure~\ref{fig:spatialattention}, object \( T_2 \) attends to nearby objects \( T_1 \) and \( T_3 \). The multi-head attention formulation is:

\begin{equation}
\label{eq:spatial-attention}
\text{MH}(Q_t^i, K_t^j, V_t^j) = \left[ \sum \text{softmax} \left( \frac{Q_t^i (K_t^j)^t}{\sqrt{d_k}} \right) V_t^j \right]_{m=1}^n,
\end{equation}

where \( Q_t^i \), \( K_t^j \), and \( V_t^j \) are linear projections for object \( i \) and its neighbors $j$ at time \( t \). After obtaining spatial information from Eq.~(8), we apply a \textbf{temporal convolution} operation on temporal edges in the spatio-temporal graph to capture dynamics over time. Trajectory data has irregular timestamps, making fixed positional encodings suboptimal. Instead, we apply 2D convolutions to capture local temporal patterns and handle missing data. This procedure complements our physics-informed design, where kinematic features (e.g., velocity, bearing) naturally encode temporal dynamics. As shown in Figure~\ref{fig:temporalattention}, temporal attention uses sliding windows (e.g., $t-1$ to $t+1$) to model stepwise dependencies. Final spatiotemporal attention is formed via multiple skip connections and layer normalization, i.e., \( \text{LayerNorm}(x + \text{Sublayer}(x)) \), to ensure stability within the transformer-based spatio-temporal graph as demonstrated by Encoder in Figure \ref{fig:Transformer}.

% \textbf{Temporal Attention:} 
% Trajectory data has irregular timestamps, making fixed positional encodings suboptimal. Instead, we apply 2D convolutions to capture local temporal patterns and handle missing data. This procedure complements our physics-informed design, where kinematic features (e.g., velocity, bearing) naturally encode temporal dynamics. As shown in Figure~\ref{fig:temporalattention}, temporal attention uses sliding windows (e.g., $t-1$ to $t+1$) to model stepwise dependencies. Final spatiotemporal attention is formed via skip connections and layer normalization: \( \text{LayerNorm}(x + \text{Sublayer}(x)) \).

\subsection{Physics-informed Decoder} 
\begin{figure}[htbp]
\centering
\includegraphics[width=0.95\columnwidth]{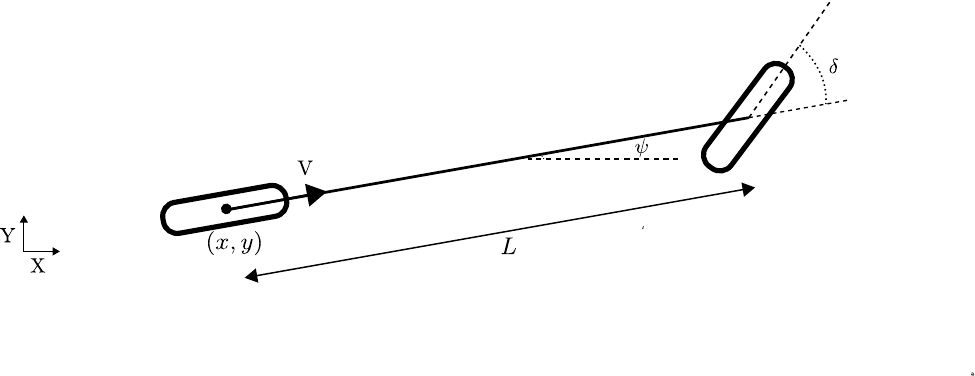}
\caption{The kinematic bicycle model}
\label{fig:kinematicModel}
\end{figure}

Data-driven models capture complex patterns but often lack physical constraints, limiting robustness in noisy or unseen scenarios. To address this, we integrate the kinematic bicycle model (KBM) \cite{polack2018guaranteeing}, a 3-DoF formulation of x-coordinate, y-coordinate, and heading angle given a constant velocity $v$.
Figure \ref{fig:kinematicModel} shows the kinematic bicycle model for a given constanct velocity $v$,

where \( x \) and \( y \) represent the coordinates of the rear wheel in the map, \( v \) denotes the velocity of the model at the rear wheel, \( L \) is the wheelbase of the vehicle, \( \delta \) is the steering angle, and \( \psi \) is the yaw angle of the vehicle. These parameters collectively define the kinematic bicycle model, which is used to describe the vehicle's motion under the given assumptions.

\begin{figure}[ht!]
    \centering
    \begin{subfigure}[b]{0.35\columnwidth}
        \centering
        \includegraphics[width=\linewidth]{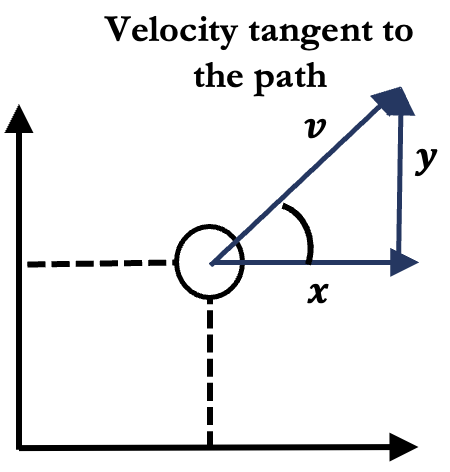} % Replace with your actual path
        \caption{2D bicycle model}
        \label{fig:KinematicConstraints}
    \end{subfigure}
    \hfill
    \begin{subfigure}[b]{0.55\columnwidth}
        \centering
        \includegraphics[width=\linewidth]{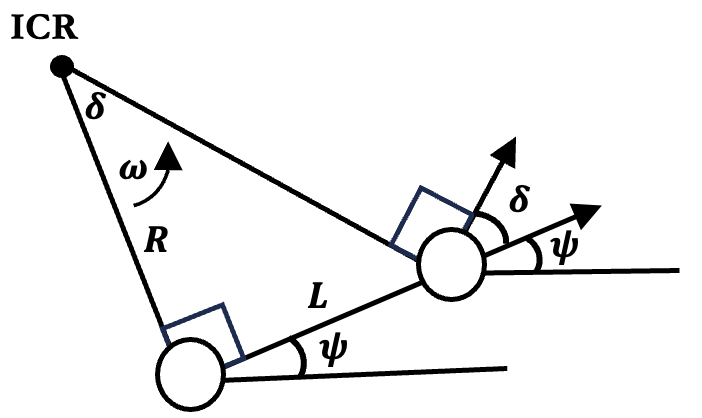} % Replace with your actual path
        \caption{Rear Axle Bicycle Model}
        \label{fig:RearAxleBicycleModel}
    \end{subfigure}
    \caption{Kinematic Bicycle Model (KBM)}
    \label{fig:simplifiedkbm}
\end{figure}

KBM assumes negligible slip, slope, and inertial effects, making it suitable for low-speed scenarios. It models vehicle motion using a simplified structure, where the wheels are lumped into a single axis, and pitch, roll, and vertical dynamics are omitted, with curvature or steering-based control inputs. In Figure \ref{fig:simplifiedkbm}, we adopt a reduced form where the vehicle slip angle (angle between the vehicle velocity and heading) is assumed to be zero \cite{ghosn2022learning,rajamani2011vehicle}, yielding a computationally efficient yet physically grounded trajectory model.

% The state of the kinematic bicycle model is defined by \( z=[x, y, \psi]^\top \), and its input is defined by \( u=[V, \delta] \). The state evolution is then defined by:
% \begin{equation}\label{kinematicBicycle.eq}
%     \dot{z} = f(z,u) = \begin{bmatrix} v \cos(\psi) & v\sin(\psi) & \frac{v \tan\delta}{L}\end{bmatrix}^\top,
% \end{equation}

Figure~\ref{fig:KinematicConstraints} shows the simplified 2D bicycle model commonly used for normal driving conditions. To ensure smooth and realistic predictions, we regularize trajectory generation  \cite{polack2017kinematic}, which is sufficient for datasets without high-speed maneuvers. The KBM models vehicle state is defined by \( z=[x, y, \psi]^\top \), and its input is defined by \( u=[v, \delta] \), where $(x, y)$ is the position, $\psi$ is the heading, and $v$ is the velocity. Substituting $\quad {x} = v \cos(\psi)$, $\quad{y} = v \sin(\psi)$, and ${\psi} = \omega$ using the rear axle formulation depicted in Figure~\ref{fig:RearAxleBicycleModel}). The model's state evolution is governed by:

\begin{equation}\label{kinematicBicycle.eq}
    \dot{z} = f(z,u) = \begin{bmatrix} v \cos(\psi) & v\sin(\psi) & \frac{v \tan\delta}{L}\end{bmatrix}^\top,
\end{equation}

where $\psi$ is the heading angle (also $\psi = \omega$), and $\omega = v / R$ is the rotation rate with turning radius \( R = L / \tan(\delta) \). Substituting yields \( \dot{\psi} = v \cdot \tan(\delta) / L \). The rate of change of heading, \(\dot{\psi}\), is driven by the curvature input \(\kappa\), defined at the front axle center. The KBM maps control inputs to state derivatives using vehicle-specific parameters like wheelbase. However, since our real-world dataset lacks such details (e.g., wheelbase, steering angle), we use curvature \(\kappa\) as the control input. This enables a vehicle-agnostic formulation of the motion model based on the state \(\mathbf{x} = (x, y, \psi, v)\):

\begin{equation}
\frac{d\mathbf{x}}{dt}
=\frac{d}{dt} \begin{bmatrix} x \\ y \\ \psi \\ v \end{bmatrix}
= \begin{bmatrix} v \cos(\psi) \\ v \sin(\psi) \\ v \kappa \\ a \end{bmatrix}
\end{equation}

\subsection{Training and Inference}
To effectively train the diffusion model, we derive its training objective by minimizing the negative log-likelihood of the model's predictive distribution under the real data expectation. This approach is consistent with other generative models and focuses on optimizing the cross-entropy between the model’s predictions and the true data distribution:

\begin{equation}
    \mathcal{L} = \mathbb{E}_{x_0 \sim q(x_0)} \left[ - \log p_{\theta} (x_0) \right]
\end{equation}

Since computing the cross-entropy directly is challenging, we introduce Jensen’s inequality to derive a variational lower bound (VLB) that facilitates optimization:

\begin{align}
\mathcal{L} &= -\mathbb{E}_{q(x_{0:T})} \log p_{\theta}(x_0) \notag \\
&\leq -\mathbb{E}_{q(x_{0:T})} \log \frac{p_{\theta}(x_{0:T})}{q(x_{1:T} \mid x_0)} \notag \\
&= \mathbb{E}_{q(x_{0:T})} \left[ \log \frac{q(x_{1:T} \mid x_0)}{p_{\theta}(x_{0:T})} \right] \notag \\
&= \mathcal{L}_{\text{VLB}}
\end{align}

This step reformulates the problem into minimizing the KL-divergence between the two distributions. Further derivation results in a decomposition of the loss into entropy and multiple KL-divergence terms:

\begin{equation}
\mathcal{L}_{\text{VLB}} = \mathbb{E}_q \left[ -\log p_\theta(x_T) + \sum_{t=1}^{T} \log \frac{q(x_t \mid x_{t-1})}{p_\theta(x_{t-1} \mid x_t)} \right]
\end{equation}

This is further decomposed as:
\begin{align}
\mathcal{L}_{\text{VLB}} &= \mathbb{E}_q \left[ - \log p_\theta(x_0 \mid x_1) + \mathrm{D_{KL}} \left( q(x_T \mid x_0) \parallel p_\theta(x_T) \right) \right. \notag \\
&\quad + \left. \sum_{t=2}^{T} \mathrm{D_{KL}} \left( q(x_{t-1} \mid x_t, x_0) \parallel p_\theta(x_{t-1} \mid x_t) \right) \right]
\end{align}

To simplify this, we assume both $q$ and $p_\theta$ are Gaussian distributions, The KL divergence between them becomes:

\begin{equation}
\mathcal{L}_{\text{VLB}} = \mathbb{E}_q \left[ \frac{1}{2} \left\| \Sigma_\theta(x_t, t)^{-1/2} \left( \tilde{\mu}_t(x_t, x_0) - \mu_\theta(x_t, t) \right) \right\|^2 \right] + C
\end{equation}

Finally, by assuming $\Sigma_\theta(x_t, t) = \sigma_t^2 \mathbf{I}$ and combining everything, we arrive at the simplified training objective:

\begin{equation}
\mathcal{L}_{\text{simple}} = \mathbb{E}_{x_0, \epsilon_t, t} \left[ \left\| \epsilon_t - \epsilon_\theta \left( \sqrt{\bar{\alpha}_t} x_0 + \sqrt{1 - \bar{\alpha}_t} \epsilon_t, t \right) \right\|^2 \right]
\end{equation}

This form clearly shows that diffusion model training amounts to denoising: learning the mean squared error between real noise and predicted noise. Once the diffusion model $\epsilon_\theta$ is trained, we are able to generate $x_{t-1}$ from $x_t$ according to Eq.~(3), iteratively, until the original sample $x_0$ is reconstructed. Then, anomaly detection is performed by comparing $x_0$ with the reconstructed trajectory $\hat{x}_0$. From the reparameterization and Eq.~(16), the sampling process to obtain $x_{t-1}$ is given by:

\begin{equation}
x_{t-1} = \sqrt{\bar{\alpha}_{t-1}} \left( \frac{x_t - \sqrt{1 - \bar{\alpha}_t} \epsilon_\theta(x_t, t)}{\sqrt{\bar{\alpha}_t}} \right)
+ \sqrt{1 - \bar{\alpha}_{t-1} - \sigma_t^2} \cdot \epsilon_\theta(x_t, t) + \sigma_t \epsilon_t,
\label{eq:sampling}
\end{equation}

where $\epsilon_t \sim \mathcal{N}(0, \mathbf{I})$ and $\epsilon_\theta$ is the trained neural network. To control the stochasticity of the sampling, we introduce a hyperparameter $\eta$ and define $\sigma_t(\eta)^2 = \eta \cdot \tilde{\beta}_t$, where $\eta \in \mathbb{R}^+$.

\textbf{Physics-based Regularization:} To guide learning with physical plausibility, we incorporate the kinematic bicycle model (KBM) as a prior within the diffusion framework. The overall loss combines the variational lower bound, reconstruction error, and a physics loss that penalizes deviations from KBM dynamics, high acceleration, and sharp curvature, promoting smooth and realistic trajectories.

As shown in Eq.~\ref{eq:physics}, the physics loss captures violations of KBM differential constraints and acts as a regularizer against physically implausible motion and enforcing low-level kinematic constraints that reflect geometric and control limitations. 

\begin{align}
\label{eq:physics}
\mathcal{L}_{\text{Phy}} &= w_1\left\|\frac{d\hat{x}}{dt} - \hat{v} \cos\hat{\psi} \right\|^2 
+ w_2\left\|\frac{d\hat{y}}{dt} - \hat{v} \sin\hat{\psi} \right\|^2 \nonumber \\
&\quad + w_3\left\|\frac{d\hat{\theta}}{dt} - \hat{v}\hat{\kappa} \right\|^2 
+ w_4\left\|\frac{d\hat{v}}{dt} - \hat{a} \right\|^2
\end{align}

\textbf{Reconstruction-based Regularization:} To detect anomalous trajectories, we monitor the reconstruction error between predicted and ground truth trajectories. Given a reconstructed trajectory \(\hat{\vx}_0\), the reconstruction error is defined as \(E_\Delta = \| \vx_0 - \hat{\vx}_0 \|^2\), where \(\vx_0\) is initialized with \(\vx_T\). The reconstruction loss used during training captures the cumulative squared error across all trajectory points:

\begin{equation}
\label{eq:reconstruction}
\mathcal{L}_{\text{Rec}} = \sum_{t=1}^{T} \left\| \begin{pmatrix} \hat{x}_t \\ \hat{y}_t \end{pmatrix} - \begin{pmatrix} x_t \\ y_t \end{pmatrix} \right\|^2
\end{equation}

The overall training objective combines variational, reconstruction, and physics-based losses:

\begin{equation}
\label{eq:total_loss}
\mathcal{L}_{\text{Pi-DPM}} = \gamma_1 \mathcal{L}_{\text{VLB}} + \gamma_2 \mathcal{L}_{\text{Rec}} + \gamma_3 \mathcal{L}_{\text{Phy}}
\end{equation}

\section{Experiment Evaluation} \label{sec:Experiment}
\begin{figure}[ht]
    \centering
    \includegraphics[width=\textwidth]{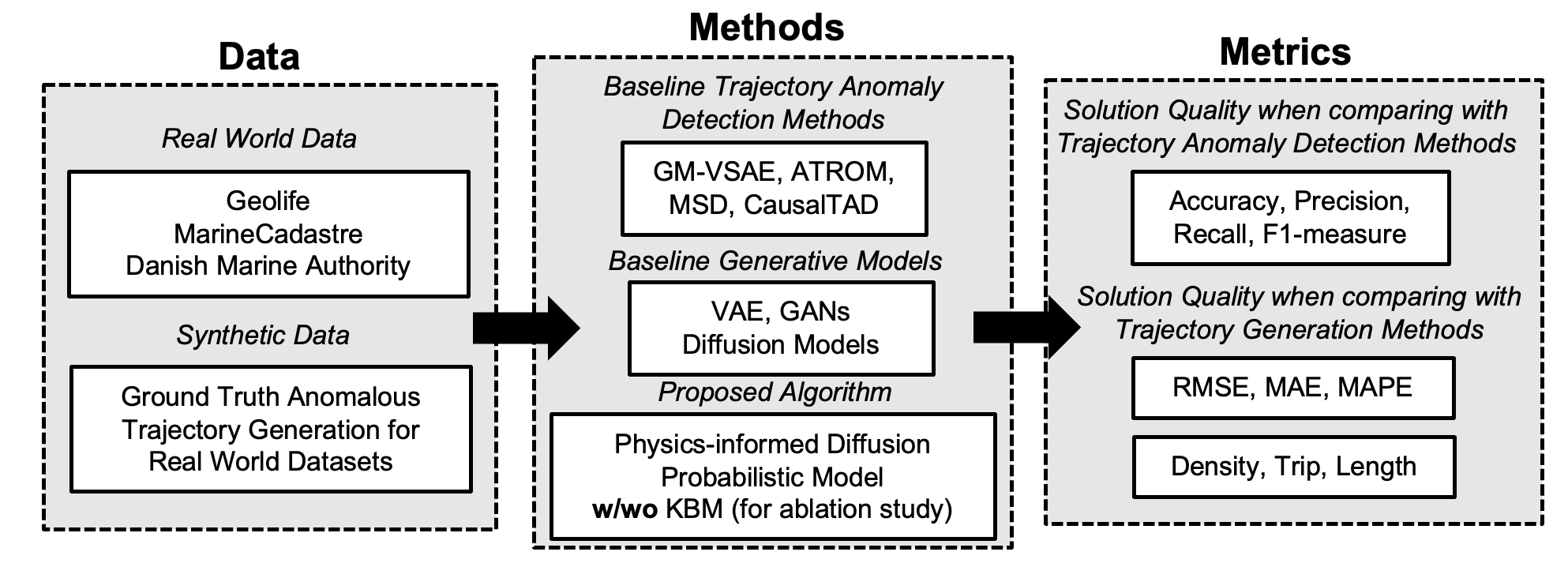}
    \caption{Experiment Design}
    \label{fig:ExperimentDesign}
\end{figure}

We conducted comprehensive experiments on real-world data, comparing Pi-DPM against state-of-the-art trajectory generative models and anomaly detection methods. We also performed ablation studies on the proposed method (e.g., Pi-DPM with and without KBM) and a sensitivity analysis of the effect of the reconstruction threshold $\lambda$. Figure \ref{fig:ExperimentDesign} illustrates the overall experiment design. The experiment goals, based on five research questions (RQs), are as follows:
\vspace{-0.5em}
\begin{itemize}\setlength\itemsep{0pt}
\item \textbf{RQ1}: How does the performance of the proposed method compare with other SOTA anomaly detection methods?
\item \textbf{RQ2}: How does the performance of the proposed method with other SOTA anomaly detection methods while considering ablation study?
\item \textbf{RQ3}: How does the proposed method compare with other anomaly detection methods as the $\lambda$ increases?
\item \textbf{RQ4}: How is the proposed method's generalizability affected when trained and tested on different datasets?
\item \textbf{RQ5}: How does the performance of the proposed method compare with trajectory generation approaches in terms of prediction accuracy?
\end{itemize}

\subsection{Experimental Design}
\textbf{Dataset:} We evaluated our Pi-DPM data model from three real-world datasets spanning diverse spatial and temporal domains, namely: (1) Geolife~\cite{zheng2010geolife}, (2) MarineCadastre~\cite{aisdataUS}, and (3) Danish Maritime Authority~\cite{aisdataDansih}. These datasets vary significantly in scale and geographic coverage. For example, MarineCadastre includes over 1.1 million maritime trajectories within a global bounding box \([-180^\circ, 180^\circ], [-90^\circ, 90^\circ]\), while the Danish Maritime dataset contains 132{,}135 coastal vessel trajectories within a bounded region around Denmark. In contrast, the Geolife dataset offers 17{,}621 urban vehicle trajectories collected within a narrow spatial range in Beijing \([116.25^\circ, 116.55^\circ], [39.75^\circ, 40.05^\circ]\). The datasets also differ in trajectory granularity, with average durations ranging from 32.76 minutes (Geolife) to 394.4 minutes (MarineCadastre), and average distances spanning from 5.91 km to over 559 km.

\textbf{Synthetic Data Generation:} Given the scarcity of annotated anomalous trajectory data across all three datasets, we synthetically injected context-aware anomalies using first- and second-order kinematic derivatives, such as abrupt speed changes, bearing discontinuities, acceleration spikes, and high-jerk events. These injected anomalies emulate realistic irregularities including unauthorized turns, erratic motion, and violations of navigational norms in the Geolife data. Since velocity is not explicitly recorded, speed-over-ground (SOG) and bearing were estimated using numerical differentiation. To ensure physical plausibility, we restricted evaluation to well-aligned road segments, avoiding inconsistencies with the urban road topology. In addition, anomalies in trajectory data can also be introduced in two primary ways: (1) by altering spatial positions (i.e., x-y coordinates) or (2) by modifying temporal characteristics (i.e., time intervals). In this paper, we focused on temporal distortions by reducing the time interval between consecutive points, which consequently induced perturbations in the computed speed (typically within a $\pm50\%$ range). To further simulate anomalies, we introduced sudden bearing shifts (\textgreater90° within short intervals), and gradual trajectory drift. Each dataset was subsequently balanced to contain \textbf{5\%,  10\%, and 20\%} anomalous trajectories. These perturbations were carefully designed to reflect deceptive behaviors observed in real-world maritime scenarios, such as the Cathay Phoenix case illustrated in Figure~2 (Section~1). To ensure a diverse and representative anomaly distribution, we systematically varied both the type and severity of the injected perturbations.

\begin{table}[h]
\caption{Hyperparameter Settings for Pi-DPM}\label{tab:paraset}
\centering
\begin{tabular}{lcrr} 
\toprule
Hyperparameter & & Setting Value & Reference Range  \\ 
\cmidrule(lr){1-4}
Diffusion Steps & & 1000  & 200 $\sim$ 500\\
Skip Steps & & 4 & 1 $\sim$ 8\\
Guidance Scale & & 2.5 & 0.5 $\sim$ 8\\
$\beta$ (linear) & & 0.0002 $\sim$ 0.04 & -- \\
Batch Size & & 512 & $\ge$ 128 \\
Sampling Blocks & & 3 &  $\ge$ 2 \\
ResNet Blocks & & 3 & $\ge$ 2\\
Input Length & & 180 & 100 $\sim$ 200\\
\bottomrule
\end{tabular}
\end{table}

\begin{table}[t]
  \scriptsize
  \setlength{\tabcolsep}{3pt}   % narrower column padding
  \caption{Pi\mbox{-}DPM with kinematic constraints detects anomalous trajectories more accurately than baselines (5\% anomalies).}
  \centering
  \resizebox{\linewidth}{!}{%
    \begin{tabular}{lcccc|cccc|cccc}
      \toprule
      \multirow{2}{*}{Methods} &
      \multicolumn{4}{c}{Geolife} &
      \multicolumn{4}{c}{MarineCadastre} &
      \multicolumn{4}{c}{Danish Maritime Authority} \\
      \cmidrule(lr){2-5}\cmidrule(lr){6-9}\cmidrule(lr){10-13}
      & Acc.\,$\uparrow$ & Prec.\,$\uparrow$ & Rec.\,$\uparrow$ & F1\,$\uparrow$ &
        Acc.\,$\uparrow$ & Prec.\,$\uparrow$ & Rec.\,$\uparrow$ & F1\,$\uparrow$ &
        Acc.\,$\uparrow$ & Prec.\,$\uparrow$ & Rec.\,$\uparrow$ & F1\,$\uparrow$ \\
      \midrule
      iBAT \cite{zhang2011ibat}      & 0.62 & 0.60 & 0.62 & 0.61 & 0.61 & 0.59 & 0.61 & 0.60 & 0.62 & 0.60 & 0.62 & 0.61 \\
      iBOAT \cite{chen2013iboat}     & 0.67 & 0.65 & 0.67 & 0.66 & 0.66 & 0.64 & 0.66 & 0.65 & 0.67 & 0.65 & 0.67 & 0.66 \\
      GM-VSAE \cite{liu2020online}   & 0.72 & 0.70 & 0.72 & 0.71 & 0.71 & 0.69 & 0.71 & 0.70 & 0.72 & 0.70 & 0.72 & 0.71 \\
      DeepTEA \cite{han2022deeptea}  & 0.77 & 0.75 & 0.77 & 0.76 & 0.76 & 0.74 & 0.76 & 0.75 & 0.77 & 0.75 & 0.77 & 0.76 \\
      ATROM \cite{gao2023open}       & 0.82 & 0.80 & 0.82 & 0.81 & 0.81 & 0.79 & 0.81 & 0.80 & 0.82 & 0.80 & 0.82 & 0.81 \\
      CausalTAD \cite{li2024causaltad} & 0.87 & 0.85 & 0.87 & 0.86 & 0.86 & 0.84 & 0.86 & 0.85 & 0.87 & 0.85 & 0.87 & 0.86 \\
      MSD-OATD \cite{wang2024multi}  & 0.90 & 0.88 & 0.90 & 0.89 & 0.89 & 0.87 & 0.89 & 0.88 & 0.90 & 0.88 & 0.90 & 0.89 \\
      LM-TAD \cite{mbuya2024trajectory} & 0.92 & 0.90 & 0.92 & 0.91 & 0.91 & 0.89 & 0.91 & 0.90 & 0.92 & 0.90 & 0.92 & 0.91 \\
      \midrule
      Pi-DPM w/o KBM (Ours) & 0.94 & 0.93 & 0.94 & 0.94 & 0.93 & 0.92 & 0.93 & 0.93 & 0.94 & 0.93 & 0.94 & 0.94 \\
      Pi-DPM w/o CIE (Ours) & 0.96 & 0.95 & 0.96 & 0.96 & 0.95 & 0.94 & 0.95 & 0.95 & 0.96 & 0.95 & 0.96 & 0.96 \\
      \textbf{Pi-DPM (Ours)} & \textbf{0.98} & \textbf{0.97} & \textbf{0.99} & \textbf{0.98} &
        \textbf{0.97} & \textbf{0.96} & \textbf{0.98} & \textbf{0.97} &
        \textbf{0.98} & \textbf{0.97} & \textbf{0.99} & \textbf{0.98} \\
      \bottomrule
    \end{tabular}}
  \caption*{\textbf{Bold} indicates the best performance; $\uparrow$ means higher is better.}
  \label{tab:acc_prf1_comp}
  \vspace{-4mm}
\end{table}

\textbf{Experimental Setup:} We implemented the Pi-DPM framework using PyTorch 2.4 and Python 3.12, executing all experiments on a single NVIDIA A100 40GB GPU. The test set was generated by selecting 20\% and remaining 80\% trips were split randomly into 60\% training and 20\% validation subsets over ten trials. We used a fixed number of diffusion steps \(\Tau = 1000\), consistent with prior work~\cite{ho2016generative,sohl2015deep,song2019generative}, and adopted a linear noise schedule with \(\beta_1 = 10^{-4}\) to \(\beta_T = 0.02\), ensuring a low signal-to-noise ratio at \(\vx_T\) while preserving alignment between the forward and reverse processes. Our implementation is available in the official github repository\footnote{\href{https://github.com/arunshar/Physics-Informed-Diffusion-Probabistic-Model}{https://github.com/arunshar/Physics-Informed-Diffusion-Probabistic-Model}} and hyperparameter ranges are reported in Table \ref{tab:paraset}.

\textbf{Evaluation Metrics:} To comprehensively assess the performance of learning-based methods for anomaly detection such as \cite{wang2024multi,mbuya2024trajectory,li2024causaltad,gao2023open}, we evaluated their effectiveness using a suite of standard metrics: Accuracy, Precision, Recall, and F-measure. These metrics provide a holistic view of model performance, capturing both the correctness of predictions and the balance between false positives and false negatives. We further compared these methods with a trajectory generation approach, where the primary objective was to synthesize trajectories that closely resemble real-world movement patterns, thereby supporting downstream applications. To quantify the similarity between generated and real trajectories, we adopted established methodologies \cite{du2023ldptrace, zhu2023difftraj}, namely Root Mean Square Error (RMSE), Mean Absolute Error (MAE), and Mean Average Precision Error (MAPE), for both comparison and ablation studies. More details on these evaluation metrics such as equations can be found in the Appendix \ref{sec:metrics}.

\textbf{Baseline methods:} We first compared our approach with established online trajectory-based models, such as iBAT \cite{zhang2011ibat} and iBOAT \cite{chen2013iboat}, which detect anomalies by identifying trajectories sharing the same source-destination pair as the reference trajectory and leveraging the concept of degree of isolation. Learning-based methods, in contrast, extract features of normal routes to compute anomaly scores for target trajectories. For instance, GM-VSAE \cite{liu2020online} employs a Gaussian mixture model to represent trajectory features in latent space, while ATROM \cite{gao2023open}, a probabilistic metric learning model, focuses on identifying the specific type of anomaly a trajectory exhibits. More advanced methods, such as DeepTEA \cite{han2022deeptea} and CausalTAD \cite{li2024causaltad}, integrate traffic patterns with time-dependent trajectory patterns, whereas MSD-OATD \cite{wang2024multi} and LM-TAD \cite{mbuya2024trajectory} incorporate spatiotemporal dependencies in asynchronous settings to a certain extent. 

We further evaluated Pi-DPM against a series of baseline models, including three representative generative models: VAE \cite{xia2018deeprailway}, TrajGAN \cite{liu2018trajgans}, and DP-TrajGAN \cite{zhang2023dp}. Additionally, we considered variants based on the diffusion model approach, namely DiffWave \cite{kong2021diffwave}, which builds on the WaveNet architecture, as well as DiffTraj \cite{zhu2023difftraj} and ControlTraj \cite{zhu2024controltraj}. These models were selected for their proven ability to capture complex temporal dependencies in trajectory data, representing a spectrum of approaches—from latent variable models to sequential GANs and denoising diffusion, ensuring a comprehensive comparison across modeling paradigms. Detailed are provided in Appendix \ref{app:baseline} and Appendix \ref{app:baseline2}.

\subsection{RQ1: Anomaly Detection} We assessed anomaly detection performance using accuracy, precision, recall, and F1-score, comparing Pi-DPM to baselines iBAT \cite{zhang2011ibat}, iBOAT \cite{chen2013iboat}, GM-VSAE \cite{liu2020online}, DeepTEA \cite{han2022deeptea}, ATROM \cite{gao2023open}, CausalTAD \cite{li2024causaltad}, MSD-OATD \cite{wang2024multi}, and LM-TAD \cite{mbuya2024trajectory}. As shown in Table~\ref{tab:acc_prf1_comp}, Pi-DPM with the Kinematic Bicycle Model (KBM) consistently outperforms all baselines. On \textbf{Geolife}, it achieves 0.98 accuracy and F1, surpassing CausalTAD (0.86 F1) and LM-TAD (0.91 F1), with similar gains on \textbf{MarineCadastre} (0.97 accuracy, 0.97 F1) and \textbf{Danish Maritime Authority} (0.98 accuracy, 0.99 recall). Among its variants, Pi-DPM w/o KBM yields the lowest performance (e.g., 0.94 F1 on Geolife), followed by Pi-DPM w/o CIE (0.96 F1 on Geolife), while Pi-DPM demonstrates consistent improvements, confirming the utility of incorporating physics-informed motion constraints. These results, based on 5 percent injected anomalies, highlight Pi-DPM’s robustness in detecting subtle deviations. Additional results for 10 percent and 20 percent anomaly settings show similar trends in Tables~\ref{tab:acc_prf1_10} and~\ref{tab:acc_prf1_20} respectively.

\begin{table}[t]
  \scriptsize                    % more compact than \small
  \setlength{\tabcolsep}{3pt}    % cut horizontal padding
  \caption{Pi\mbox{-}DPM with a kinematic bicycle model (KBM) yields the lowest prediction errors.}
  \centering
  \resizebox{\linewidth}{!}{%
    \begin{tabular}{lccc|ccc|ccc}
      \toprule
      \multirow{2}{*}{Methods} &
      \multicolumn{3}{c}{Geolife \cite{zheng2010geolife}} &
      \multicolumn{3}{c}{MarineCadastre \cite{aisdataUS}} &
      \multicolumn{3}{c}{Danish Maritime Authority \cite{aisdataDansih}} \\
      \cmidrule(lr){2-4}\cmidrule(lr){5-7}\cmidrule(lr){8-10}
      & RMSE$\downarrow$ & MAE$\downarrow$ & MAPE$\downarrow$
      & RMSE$\downarrow$ & MAE$\downarrow$ & MAPE$\downarrow$
      & RMSE$\downarrow$ & MAE$\downarrow$ & MAPE$\downarrow$ \\
      \midrule
      VAE \cite{xia2018deeprailway}          & 247.18 & 333.30 & 74.3 & 313.51 & 309.14 & 77.2 & 239.26 & 342.34 & 73.7 \\
      TrajGAN \cite{rao2020lstm}             & 216.44 & 336.23 & 69.8 & 251.25 & 269.85 & 72.7 & 213.76 & 345.63 & 71.3 \\
      DP-TrajGAN \cite{zhang2023dp}          & 208.02 & 284.26 & 66.2 & 258.69 & 232.87 & 68.2 & 207.12 & 283.68 & 66.9 \\
      Diffwave \cite{kong2021diffwave}       & 250.68 & 333.53 & 74.7 & 313.79 & 309.60 & 77.9 & 261.56 & 343.46 & 75.3 \\
      DiffTraj \cite{zhu2023difftraj}        & 225.34 & 345.13 & 72.3 & 252.73 & 272.38 & 73.1 & 215.21 & 364.66 & 71.7 \\
      ControlTraj \cite{zhu2024controltraj}  & 211.50 & 284.85 & 69.1 & 261.87 & 235.45 & 68.5 & 208.53 & 298.60 & 69.3 \\
      \midrule
      Pi\mbox{-}DPM w/o KBM (Ours) & 195.00 & 205.00 & 66.0 & 240.00 & 233.00 & 71.0 & 212.00 & 255.00 & 62.0 \\
      Pi\mbox{-}DPM w/o CIE (Ours) & 165.00 & 180.00 & 63.5 & 230.00 & 231.00 & 60.0 & 190.00 & 230.00 & 60.5 \\
      \textbf{Pi\mbox{-}DPM (Ours)} &
        \textbf{143.33} & \textbf{160.70} & \textbf{61.3} &
        \textbf{224.15} & \textbf{229.51} & \textbf{57.0} &
        \textbf{179.54} & \textbf{216.29} & \textbf{59.7} \\
      \bottomrule
    \end{tabular}}
  \caption*{\textbf{Bold} = best; $\downarrow$ indicates lower is better.}
  \label{tab:comp2}
  \vspace{-4mm}
\end{table}

\subsection{RQ2 Abalation Study} We conduct an ablation study on Geolife, MarineCadastre, and Danish Maritime Authority to assess the effectiveness of the two modules of Pi-DPM. We thus compare Pi-DPM with two variants:

\begin{itemize}[itemsep=0pt] % Set itemsep to 0pt to reduce space between items
    \item \textbf{Pi-DPM w/o CIE:} This variant does not uses Context-informed Encoder (CIE) to capture spatiotemporal dependencies for training but leverage physics-based regularizer to generate physically plausible trajectories.
    \item \textbf{Pi-DPM w/o KBM:} This variant only uses spatiotemporal dependencies and disregards Kinematic Bicycle Model (KBM) for training to detects spatial anomalous trajectories.
    \item \textbf{Pi-DPM:} This variant uses both spatiotemporal dependencies and physics-based regularizer (KBM+CIE) to generate physically plausible trajectories.
\end{itemize}

Pi-DPM outperforms both variants in our ablation study, highlighting the importance of integrating spatiotemporal dependencies and a physics-based regularizer. We observe that Pi-DPM w/o KBM, which relies solely on spatiotemporal dependencies, performs the worst, indicating the limitation of neglecting kinematic constraints in detecting physically plausible anomalies. Pi-DPM w/o CIE, which leverages a physics-based regularizer but omits spatiotemporal dependencies, shows improved performance over Pi-DPM w/o KBM, yet still falls short of the full model. This suggests that while physics-based regularization aids in generating realistic trajectories, spatiotemporal dependencies are crucial for capturing complex mobility patterns. Both variants yield consistent results across the Geolife, MarineCadastre, and Danish Maritime Authority datasets, but Pi-DPM consistently achieves higher accuracy.

\subsection{RQ3 Sensitivity Analysis} 
\begin{figure}[ht!]
    \centering
    \begin{subfigure}[b]{0.49\columnwidth}
        \centering
        \includegraphics[width=\linewidth]{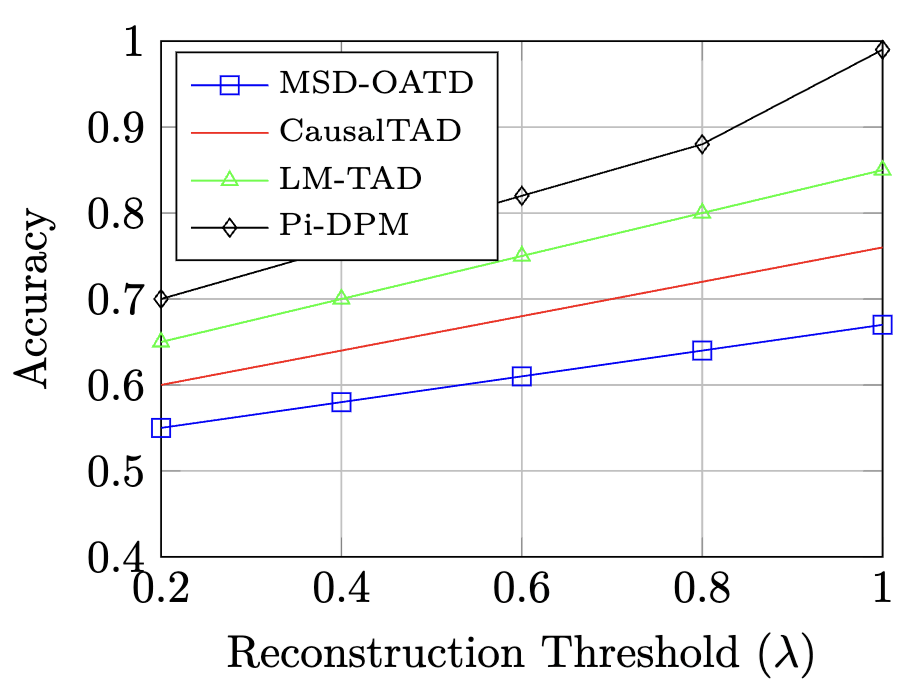}
        \caption{Accuracy}
        \label{fig:accuracy}
    \end{subfigure}
    \hfill
    \begin{subfigure}[b]{0.49\columnwidth}
        \centering
        \includegraphics[width=\linewidth]{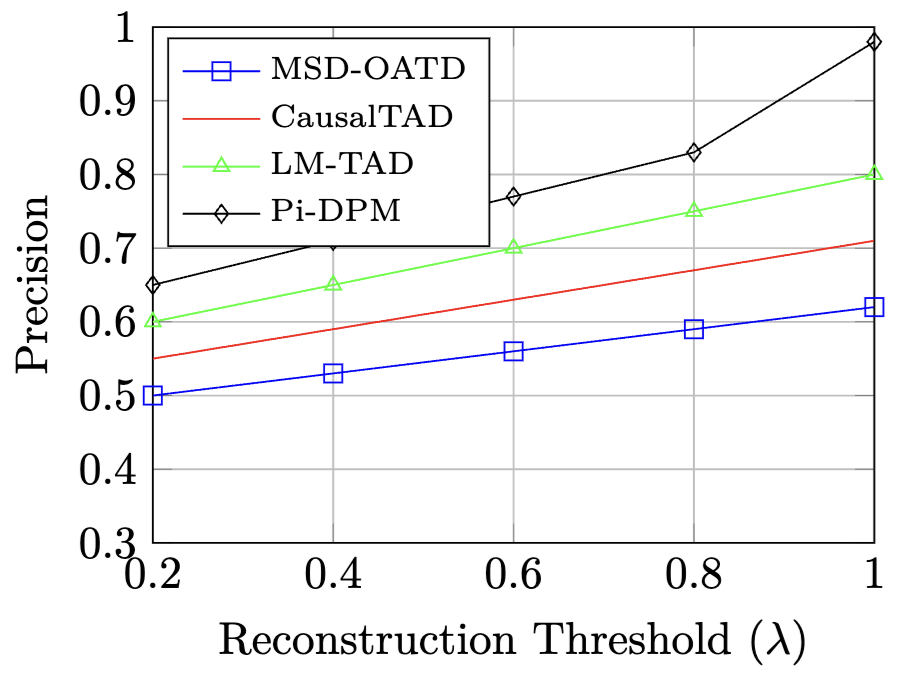}
        \caption{Precision}
        \label{fig:precision}
    \end{subfigure}

    \vskip\baselineskip

    \begin{subfigure}[b]{0.49\columnwidth}
        \centering
        \includegraphics[width=\linewidth]{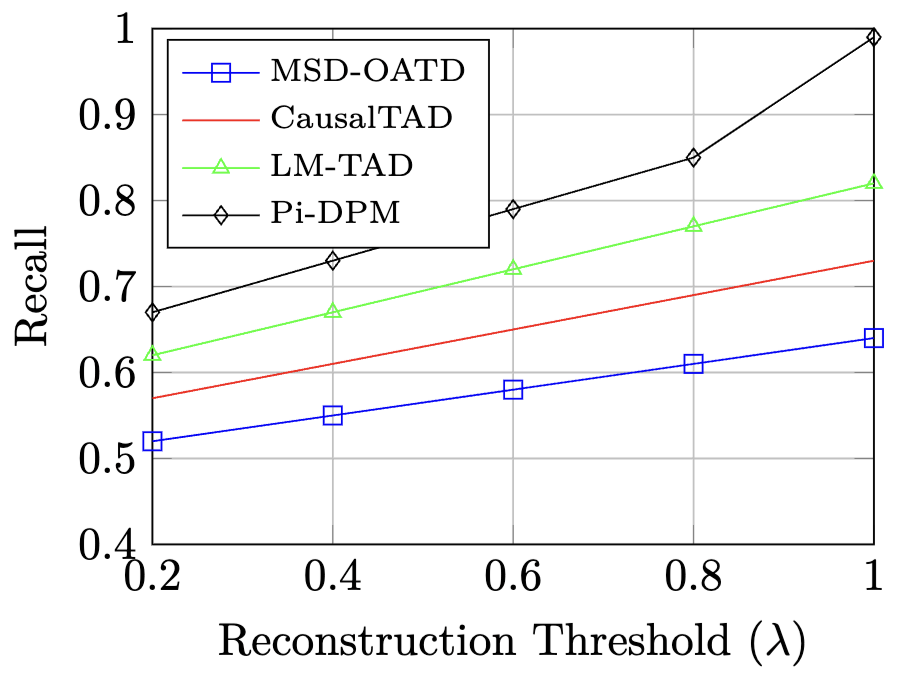}
        \caption{Recall}
        \label{fig:recall}
    \end{subfigure}
    \hfill
    \begin{subfigure}[b]{0.49\columnwidth}
        \centering
        \includegraphics[width=\linewidth]{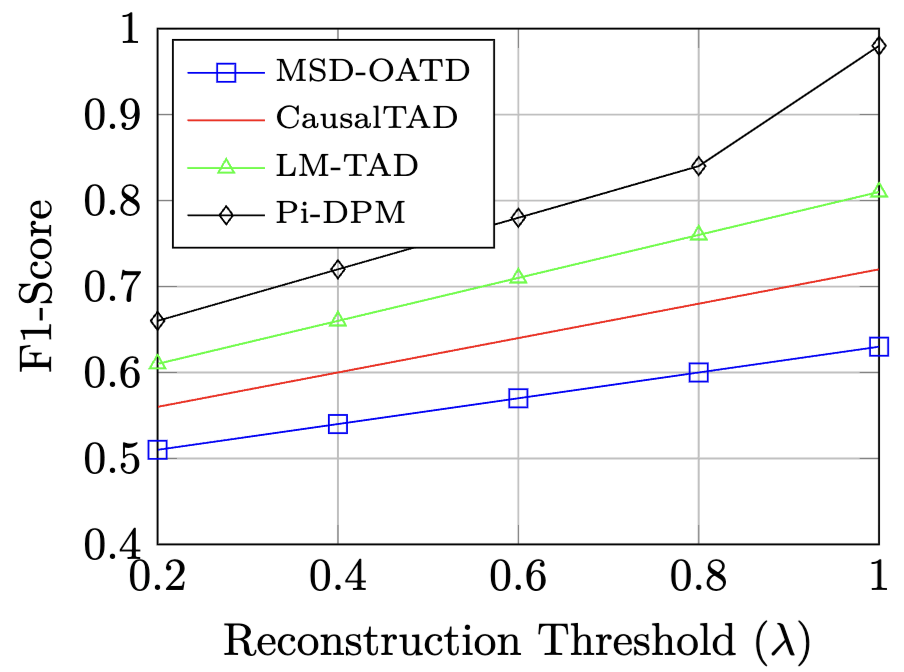}
        \caption{F-1 Score}
        \label{fig:f1score}
    \end{subfigure}

    \caption{Effect of increases in the $\lambda$ threshold}
    \label{fig:trajectory_models}
\end{figure}

We assessed the effect of the reconstruction threshold lambda on model performance by varying lambda from 0.2 to 1.0, comparing Pi-DPM with MSD-OATD, CausalTAD, and LM-TAD. Figure~\ref{fig:trajectory_models} shows all models improve with increasing lambda, but Pi-DPM consistently outperforms others across all metrics. Notably, Pi-DPM’s accuracy and F1-score rise sharply to 0.99 and 0.98 at lambda = 1.0, while MSD-OATD, CausalTAD, and LM-TAD reach only 0.67, 0.76, and 0.85 in accuracy, and 0.63, 0.72, and 0.81 in F1-score, respectively. Pi-DPM also excels in precision and recall, hitting 0.98 and 0.99. This underscores Pi-DPM’s robustness and superior anomaly detection capability using reconstruction-based criteria.

\subsection{RQ4: Transfer Learning} 
\begin{figure}[ht]
  \centering
  %------------ First subfigure (now Maritime) -------------
  \begin{subfigure}[t]{0.48\linewidth}
    \centering
    \includegraphics[width=\linewidth]{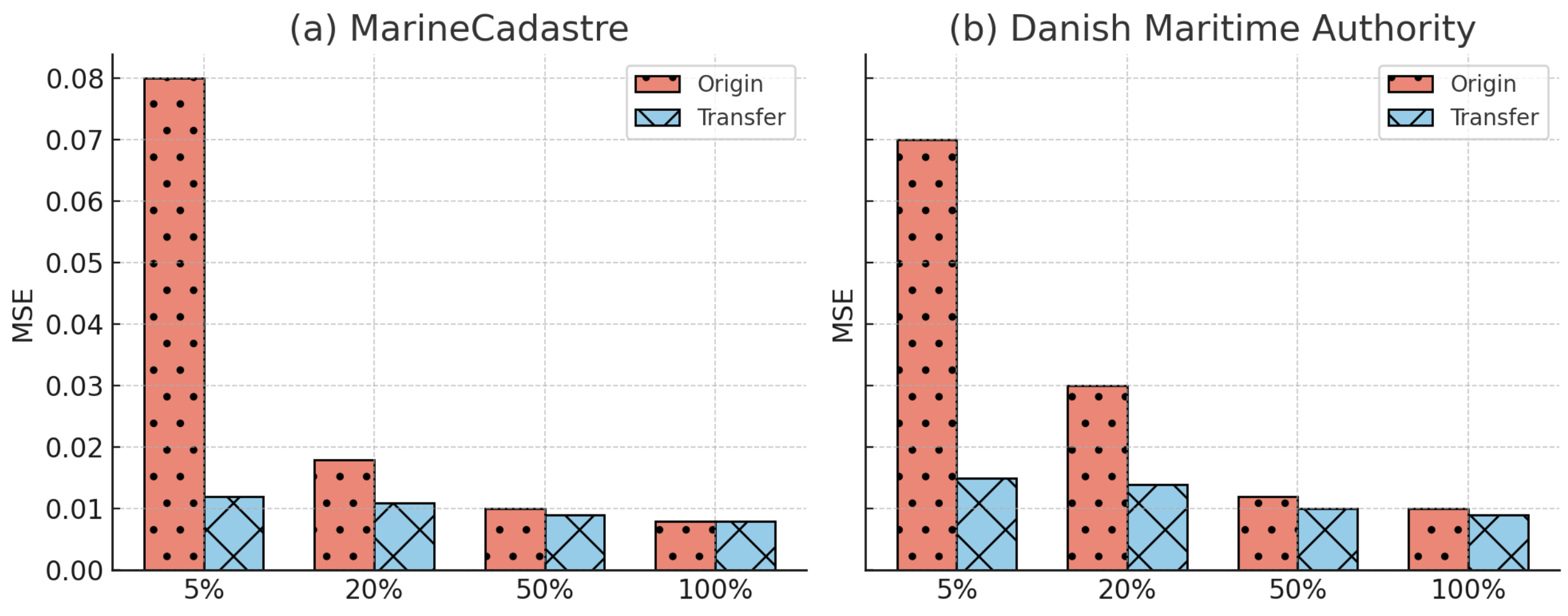}
    \caption{Transfer Learning in Maritime Domain}
    \label{fig:Transfer1}
  \end{subfigure}
  \hfill
  %------------ Second subfigure (now Across Domains) -------
  \begin{subfigure}[t]{0.48\linewidth}
    \centering
    \includegraphics[width=\linewidth]{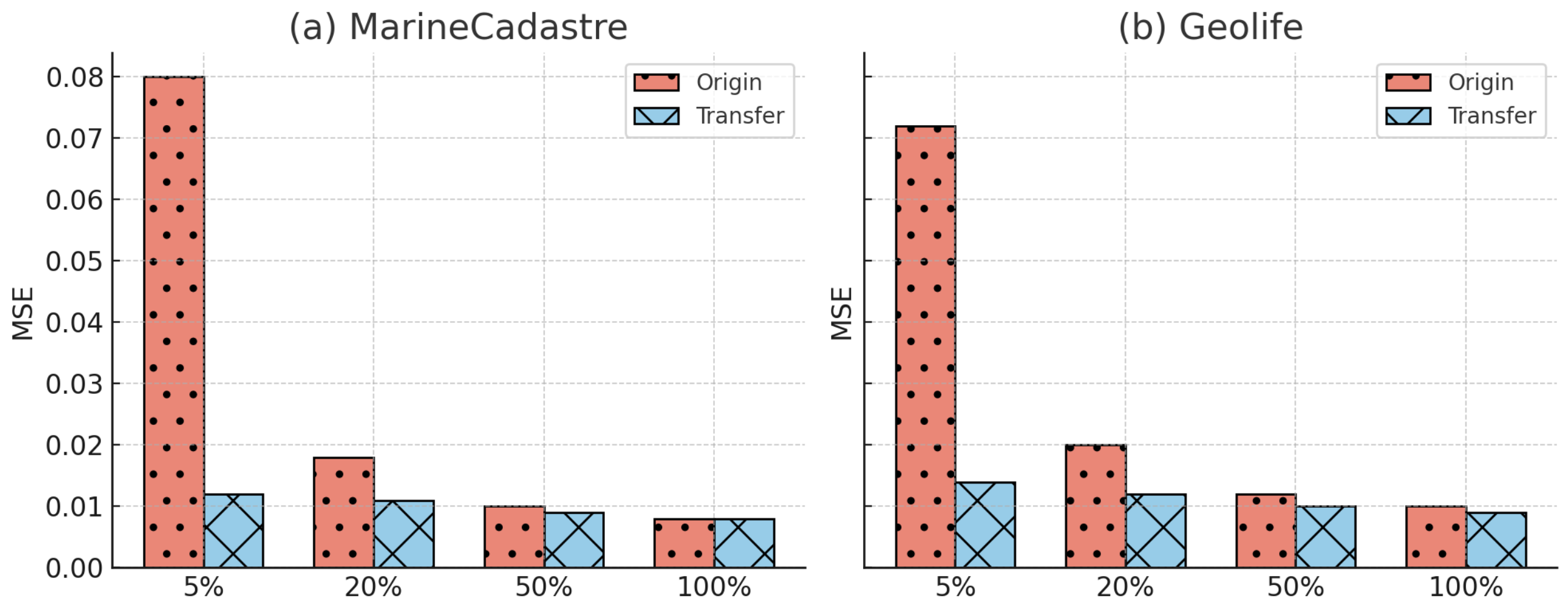}
    \caption{Transfer Learning across Domains}
    \label{fig:Transfer2}
  \end{subfigure}
  \caption{Illustrations of transfer learning scenarios.}
  \label{fig:TransferSubfigs}
\end{figure}

To evaluate the generalizability of Pi-DPM across spatial-temporal domains, we conducted transfer learning experiments by pre-training the model on one maritime region and fine-tuning it on another region using varying proportions of target domain data ($5\% \sim 100\%$). We compared this against models trained from scratch on the same data proportions to assess the extent of knowledge transfer. Figure~\ref{fig:Transfer1} shows results for transfer across maritime domains. Even with as little as 5\% target data, the transferred model achieves significantly lower reconstruction error than training from scratch. As the data percentage increases, the performance gap narrows, confirming Pi-DPM’s adaptability under limited supervision.

We further analyzed cross-domain transfer from MarineCadastre to Geolife. Figure~\ref{fig:Transfer2} shows that, despite the domain shift, the model maintains strong performance with minimal degradation, indicating that Pi-DPM effectively captures transferable trajectory structures. These findings highlight the framework’s robustness and its potential for rapid adaptation in diverse real-world scenarios such as mobility analytics, maritime surveillance, and urban planning.

% \begin{figure}[ht]
%     \centering
%     \includegraphics[width=0.38\textwidth]{./Figures/Transfer Learning 2.jpeg}
%     \caption{Transfer Learning across Domains}
%     \label{fig:Transfer2}
% \end{figure}

\subsection{RQ5: Trajectory Reconstruction}\label{sec:rq5}

As summarized in Table~\ref{tab:comp2}, the full Pi-DPM, which incorporates both the kinematic bicycle model (KBM) prior and the cross-interval encoder (CIE), attains the lowest reconstruction errors—RMSE, MAE, and MAPE—on all three benchmarks, demonstrating the value of embedding physics-informed knowledge in a diffusion backbone.  
On Geolife, RMSE drops from 247.18 for the latent-variable VAE to 143.33, with comparable improvements in MarineCadastre and DMA. Traditional baselines (VAE, TrajGAN, DP-TrajGAN) struggle because they emphasize spatial distributions while modeling temporal correlations only weakly; consequently they exhibit the highest MAPE ($\geq$ 66 \% on every dataset). Diffusion baselines (DiffWave, DiffTraj, ControlTraj) refine trajectories step by step from noise and therefore capture stochastic dynamics better, but they remain motion-agnostic—for example, ControlTraj lowers Geolife RMSE to 211.50 yet still exceeds our method by 68.2 units. Ablation studies highlight that both inductive components are essential. Removing KBM raises RMSE from 143.33 to 195.00 and inflates MAPE from 61.3 \% to 66.0 \% on Geolife, while omitting CIE (keeping KBM) further degrades the accuracy.  Similar additive effects are observed on MarineCadastre and DMA, confirming that KBM enforces kinematic feasibility and CIE captures long-range context. These results underscore the advantage of combining diffusion-based generative modeling with physics-informed priors for high-fidelity trajectory generation and highlight a promising direction for future work.

\section{Related Work}\label{related_work}
Traditional trajectory mining addresses data management, and uncertainty, as well as taxonomies \cite{zheng2015trajectory,dodge2008towards}. Trajectory synthesis splits into rule-based and data-driven methods where the former fixes speed, heading, and clustering rules, whereas the latter learns patterns directly from historical races \cite{pfoser2003generating}. Microscopic traffic simulators \cite{krajzewicz2012recent,fellendorf2010microscopic,stanford2024numosim} and their city-scale extensions \cite{zhang2019cityflow,liang2023cblab} embed car-following and lane-change heuristics \cite{aycin1999comparison} yet struggle with the intrinsic stochasticity of human mobility. In anomaly detection \cite{chandola2009anomaly}, similarity methods such as edit distance, LCS, DTW, and sliding windows \cite{chen2005robust,lin1995fast,yi1998efficient,bakalov2005efficient,chen2009design} actively shown promise in early literature of anomaly detection in trajectories, whereas simulation-based mobility generators \cite{simini2021deep} and noise-based perturbations \cite{zandbergen2014ensuring} methods still hinges on historical mobility behavior assumptions which does not capture fine-scale spatiotemporal dependencies \cite{li2024difftad,xue2021mobtcast,meloni2011modeling}. For instance, recent literature \cite{zhang2011ibat,chen2013iboat} approximate study area to a coarse-scale grid cells which limits complex spatiotemporal interactions. Other literature which do consider spatiotemporal interactions \cite{liu2020online,wang2024multi,gao2023open,li2024causaltad,han2022deeptea} via leveraging generative models assumes entire trip or routes which are either spatially (distant trajectories) or temporally (different time of the day) asynchronous. Hence, these methods struggle to detect anomalies at a broader scale (e.g., across routes or trips), limiting their ability to capture complex mobility behaviors \cite{xue2021mobtcast,meloni2011modeling}.

Deep generative models have introduced a new paradigm for trajectory synthesis, classified into discrete and continuous methods. Discrete models generate coarse-grained outputs such as grid-cell sequences \cite{feng2020learning, long2023practical, li2024geo}, pixel-based maps \cite{cao2021generating, lin2023origin}, road segments \cite{jiang2023continuous, wang2022deep, choi2021trajgail}, or POI transitions \cite{wang2024spatiotemporal, wang2023pategail, deng2024revisiting}, favoring computational efficiency at the expense of spatial and temporal resolution. In contrast, continuous models offer finer fidelity, aligning more closely with raw GPS traces. In addition, VAEs and GANs have enabled modeling of complex trajectory distributions \cite{henke2023condtraj,xia2018deeprailway,rao2020lstm}. However, these models often convert trajectories into low-resolution representations such as grids \cite{feng2020learning,ouyang2018non,yuan2022activity} or images \cite{wang2021large,goodfellow2014generative}, which degrade spatial-temporal accuracy \cite{luca2021survey}. While effective for group-level simulation, they fall short in reconstructing fine-grained, individual trajectories, diffusion-based models \cite{ho2020,zhu2023difftraj} have recently emerged as a compelling alternative, offering continuous and high-fidelity generation capabilities suitable for anomaly detection. However, purely data-driven methods often exhibit limitations in scientific domains due to their reliance on large-scale labeled data and inability to honor physical laws, resulting in implausible outputs in fields like climate science and biology \cite{willard2022integrating,faghmous2014big,alber2019integrating}. Our proposed method addresses these gaps by embedding kinematic constraints from the bicycle model \cite{polack2017kinematic}, ensuring that generated trajectories are not only statistically coherent but also physically feasible. This integration bridges the gap between data-driven modeling and physical realism for real-world applications.

\section{Discussion}\label{sec:Discussion}
Unlike our baseline methods, which rely solely on spatial or statistical patterns, Pi-DPM captures physically infeasible behaviors that may otherwise go undetected. For example, a replayed trajectory at abnormal speed may appear legitimate to a baseline model but is flagged by ours due to physics violations. This highlights the importance of domain-aware anomalies, especially for edge cases common in maritime and aviation domains. For instance, circular patterns are most common in open waters \cite{triebert2023fake} which violates additional physics and displacement constraints can verified such patterns multiple times by maritime authorities. Another common GPS spoofing pattern is \textit{record and replay} based anomalies that mimic historical trajectories but violate physical motion constraints.

\section{Conclusion and Future Work}
\label{sec:Conclusion}
We proposed a Physics-Informed Diffusion Probabilistic Model (Pi-DPM) to capture deception-based abnormal behavior via contextual and physical parameters. We further integrated the kinematic bicycle model \cite{polack2017kinematic} with the Denoising Diffusion Probabilistic Model (DDPM) at the decoder stage, leveraging physical-parameters for better reconstruction error estimation. Experimental results demonstrate that the Pi-CDPM outperforms state-of-the-art methods, generating more physically plausible trajectories and achieving more accurate reconstruction error.

\textbf{Future Work:} We plan to investigate whether incorporating additional physical parameters and more sophisticated kinematic models based on a four-wheel assumption \cite{rajamani2011vehicle,gillespie1992fundamentals} can yield more accurate estimates of maneuvering behavior \cite{sharma2024physics,sharma2020analyzing,sharma2022abnormal,sharma2022spatiotemporal}. We also intend to apply Pi-DPM to other datasets (e.g., on-board vehicle diagnostic data \cite{li2023eco,yang2024towards}) and to integrate multi-modal auxiliary datasets \cite{ghosh2024towards,sharma2018webgiobe,kumar2015graph} via contrastive learning \cite{li2023self,chang2023contrastive} in order to evaluate its versatility and robustness. In parallel, we will work on optimizing the computational efficiency of the diffusion model to make it more suitable for real-world deployments—an aspect that is critical for maritime and public-safety applications \cite{sharma2022understanding,li2022cscd,ghosh2022towards}. Finally, we aim to develop foundation models \cite{mai2024foundation,shekhar2025towards} that support a broad range of downstream, domain-specific tasks \cite{farhadloo2024towards,farhadloo2024spatial,farhadloo2025spatially,farhadloo2022samcnet,sharma2025spatial,gupta2022mining}, including physics-informed mobility modeling, and that can be fine-tuned for diverse anomaly-detection scenarios \cite{upadhyay2016novel,upadhyay2018novel}.

\section*{Acknowledgments}{This material is based on work supported by the USDA under Grant No. 2023-67021-39829, the National Science Foundation under Grant No. 1901099, the USDOE Office of Energy Efficiency and Renewable Energy under FOA No. DE-FOA0002044. We also thank Kim Kofolt and the Spatial Computing Research Group for their valuable comments and contributions.}

\bibliographystyle{ACM-Reference-Format}
\bibliography{sample-base}

\appendix

\begin{appendix}
\clearpage

\section{Baseline Anomaly Detection Methods}\label{app:baseline}

\begin{itemize}[leftmargin=*]

\item \textbf{iBAT:} Trajectories are first embedded in a spatio-temporal feature space (speed, heading change, stop duration, etc.) and the space is recursively split by random hyper-planes, à la Isolation Forest. Rare paths are isolated after only a few splits, so a short isolation depth directly yields a high anomaly score \cite{zhang2011ibat}. 

\item \textbf{iBOAT:} iBOAT slides a fixed-length window along each route and, for every window, counts its k nearest neighbours in a large corpus of normal traffic. Windows with little neighbor support are deemed isolated and given high anomaly scores; the voyage’s final score is the maximum window score, so even a brief deviation is enough to flag the whole track \cite{chen2013iboat}.

\item \textbf{GM-VSAE:} The Gaussian-Mixture Variational Sequence AutoEncoder learns rich spatio-temporal dependencies along a route, then clusters the resulting embeddings with a Gaussian-mixture prior to map each major “normal” route type into its own region of a continuous latent space. New trajectories are quickly projected into this space; those that cannot be well reconstructed or that fall outside the high-density mixture flagged as anomalies \cite{liu2020online}.

\item \textbf{DeepTA:} DeepTEA is a deep-probabilistic, time-aware anomaly detector that learns the evolving distribution of vehicle states, allowing it to flag outliers even under highly complex traffic patterns. To enable live monitoring, a lightweight approximate version of the model trades a small amount of accuracy for the speed needed to surface abnormal behaviour in real time \cite{han2022deeptea}.

\item  \textbf{ATROM:} The Anomaly Typology Recognition and Organization Model (ATROM) introduces a structured typology to classify anomalies across five dimensions: structure, distribution, context, semantics, and temporal patterns. This domain-independent framework improves interpretability, consistency, and adaptability across diverse spatiotemporal anomaly detection scenarios \cite{gao2023open}.

\item  \textbf{CausalTAD:} CausalTAD integrates spatial clustering with temporal prediction for trajectory anomaly detection. It uses DBSCAN to form baseline clusters of normal behavior and trains a ProbSparse Transformer to model recent motion sequences. Discrepancies between predicted and clustered behavior—measured via dynamic time warping (DTW)—are used to flag anomalies. The method emphasizes causality and time-aware interpretation of deviation \cite{li2024causaltad}.

\item  \textbf{MSD-OATD:} MST-OATD is a multi-scale system for real-time trajectory anomaly detection, capturing movement patterns at several spatial and temporal granularities. An integrated ranking mechanism continually refreshes its reference set with incoming data, allowing the model to adapt to shifting traffic behaviours while preserving detection accuracy \cite{wang2024multi}.

\item \textbf{LM-TAD:} LM-TAD reinterprets a trajectory as a language-like token sequence and trains an autoregressive, causal-attention model to learn its joint probability distribution. User-specific tokens personalise the context, so points with low likelihood (high perplexity or surprisal) are flagged as anomalies; this yields state-of-the-art results on the PoL dataset and competitive performance on Porto taxis while supporting GPS, stay-point, or activity tokens. An efficient key-value cache lets the model operate online without repeatedly recomputing attention, keeping latency low \cite{mbuya2024trajectory}.
\end{itemize}

\section{Baseline Trajectory Generation Methods}\label{app:baseline2}

\begin{itemize}[leftmargin=*]
\item \textbf{Variational Autoencoder (VAE):} The VAE encodes trajectories into a lower-dimensional latent space using two convolutional layers followed by a fully connected layer. This compact representation captures key spatiotemporal features. A symmetric decoder reconstructs the trajectory using a fully connected layer followed by two deconvolutional layers. The model is trained by minimizing a combination of reconstruction loss and KL divergence, encouraging the latent space to follow a Gaussian prior \cite{xia2018deeprailway}.

\item \textbf{Trajectory GAN (TrajGAN):} TrajGAN synthesizes plausible trajectory data by training a generator–discriminator pair. The generator takes as input a noise vector concatenated with trajectory seeds, passing it through two fully connected layers followed by two convolutional layers to output synthetic trajectories. The discriminator evaluates the authenticity of the generated sequences via convolutional and dense layers. The adversarial training framework iteratively optimizes both networks to balance fidelity and diversity \cite{rao2020lstm}.

\item \textbf{DP-TrajGAN:} The DP-TrajGAN model addresses privacy preserving trajectory synthesis by integrating an enhanced LSTM-based GAN with formal differential privacy (DP) mechanisms. The generator captures trajectory dynamics, while privacy is preserved by injecting calibrated noise into the training process. To optimize the privacy–utility trade-off, a Partially Observable Markov Decision Process (POMDP) is employed to adaptively allocate the privacy budget across training iterations. This results in synthetic trajectories that maintain statistical similarity to real-world patterns while safeguarding sensitive location information \cite{zhang2023dp}.

\item \textbf{Diffwave:} Originally designed for speech synthesis, Diffwave is a generative model based on the WaveNet architecture. It comprises 16 residual layers employing bidirectional dilated convolutions and skip connections to propagate contextual information across time steps. The output is transformed via tanh and sigmoid activations into a 1D convolutional decoder. Its autoregressive structure allows it to model long-range dependencies, making it amenable to sequential data generation tasks such as trajectory synthesis \cite{kong2021diffwave}.

\item \textbf{DiffTraj:} DiffTraj leverages diffusion processes to progressively refine noise into structured trajectories. Unlike GANs, it does not rely on adversarial training. The model incorporates conditioning via learned embeddings of start and end locations, thereby improving semantic coherence with respect to origin–destination constraints. This conditioning facilitates location-aware generation without enforcing explicit topological priors \cite{zhu2023difftraj}.

\item \textbf{ControlTraj:} ControlTraj introduces a topology-constrained diffusion model for high-fidelity and geography-aware trajectory generation. It incorporates a novel road segment autoencoder to learn fine-grained embeddings of underlying street networks. These embeddings, combined with trip-level attributes, guide a modified geographic denoising UNet architecture (GeoUNet), enabling the generation of plausible trajectories conditioned on structural constraints. ControlTraj demonstrates adaptability across diverse spatial contexts and preserves geographical realism \cite{zhu2024controltraj}.
\end{itemize}

\begin{table}[t]
  \scriptsize                     % more compact than \small
  \setlength{\tabcolsep}{3pt}     % tighten horizontal padding
  \caption{Pi\mbox{-}DPM with kinematic constraints detects anomalous trajectories more accurately than baselines (10\% anomalies).}
  \centering
  \resizebox{\linewidth}{!}{%
  \begin{tabular}{lcccc|cccc|cccc}
    \toprule
    \multirow{2}{*}{Methods} &
      \multicolumn{4}{c}{Geolife} &
      \multicolumn{4}{c}{MarineCadastre} &
      \multicolumn{4}{c}{Danish Maritime Authority} \\
    \cmidrule(lr){2-5}\cmidrule(lr){6-9}\cmidrule(lr){10-13}
      & Acc.$\uparrow$ & Prec.$\uparrow$ & Rec.$\uparrow$ & F1$\uparrow$
      & Acc.$\uparrow$ & Prec.$\uparrow$ & Rec.$\uparrow$ & F1$\uparrow$
      & Acc.$\uparrow$ & Prec.$\uparrow$ & Rec.$\uparrow$ & F1$\uparrow$ \\
    \midrule
    iBAT \cite{zhang2011ibat}        & 0.60 & 0.58 & 0.60 & 0.59 & 0.59 & 0.57 & 0.59 & 0.58 & 0.60 & 0.58 & 0.60 & 0.59 \\
    iBOAT \cite{chen2013iboat}       & 0.65 & 0.63 & 0.65 & 0.64 & 0.64 & 0.62 & 0.64 & 0.63 & 0.65 & 0.63 & 0.65 & 0.64 \\
    GM-VSAE \cite{liu2020online}     & 0.70 & 0.68 & 0.70 & 0.69 & 0.69 & 0.67 & 0.69 & 0.68 & 0.70 & 0.68 & 0.70 & 0.69 \\
    DeepTEA \cite{han2022deeptea}    & 0.75 & 0.73 & 0.75 & 0.74 & 0.74 & 0.72 & 0.74 & 0.73 & 0.75 & 0.73 & 0.75 & 0.74 \\
    ATROM \cite{gao2023open}         & 0.80 & 0.78 & 0.80 & 0.79 & 0.79 & 0.77 & 0.79 & 0.78 & 0.80 & 0.78 & 0.80 & 0.79 \\
    CausalTAD \cite{li2024causaltad} & 0.85 & 0.83 & 0.85 & 0.84 & 0.84 & 0.82 & 0.84 & 0.83 & 0.85 & 0.83 & 0.85 & 0.84 \\
    MSD-OATD \cite{wang2024multi}    & 0.88 & 0.86 & 0.88 & 0.87 & 0.87 & 0.85 & 0.87 & 0.86 & 0.88 & 0.86 & 0.88 & 0.87 \\
    LM-TAD \cite{mbuya2024trajectory}& 0.90 & 0.88 & 0.90 & 0.89 & 0.89 & 0.87 & 0.89 & 0.88 & 0.90 & 0.88 & 0.90 & 0.89 \\
    \midrule
    Pi\mbox{-}DPM w/o KBM (Ours) & 0.92 & 0.91 & 0.92 & 0.92 & 0.91 & 0.90 & 0.91 & 0.91 & 0.92 & 0.91 & 0.92 & 0.92 \\
    Pi\mbox{-}DPM w/o CIE (Ours) & 0.94 & 0.93 & 0.94 & 0.94 & 0.93 & 0.92 & 0.93 & 0.93 & 0.94 & 0.93 & 0.94 & 0.94 \\
    \textbf{Pi\mbox{-}DPM (Ours)} &
      \textbf{0.96} & \textbf{0.95} & \textbf{0.97} & \textbf{0.96} &
      \textbf{0.95} & \textbf{0.94} & \textbf{0.96} & \textbf{0.95} &
      \textbf{0.96} & \textbf{0.95} & \textbf{0.97} & \textbf{0.96} \\
    \bottomrule
  \end{tabular}}
  \caption*{\textbf{Bold} = best; $\uparrow$ indicates higher is better.}
  \label{tab:acc_prf1_10}
  \vspace{-4mm}
\end{table}

\begin{table}[t]
  \scriptsize                     % smaller than \small
  \setlength{\tabcolsep}{3pt}     % tighten horizontal padding
  \caption{Pi\mbox{-}DPM with kinematic constraints detects anomalous trajectories more accurately than baselines (20\% anomalies).}
  \centering
  \resizebox{\linewidth}{!}{%
  \begin{tabular}{lcccc|cccc|cccc}
    \toprule
    \multirow{2}{*}{Methods} &
      \multicolumn{4}{c}{Geolife} &
      \multicolumn{4}{c}{MarineCadastre} &
      \multicolumn{4}{c}{Danish Maritime Authority} \\
    \cmidrule(lr){2-5}\cmidrule(lr){6-9}\cmidrule(lr){10-13}
      & Acc.$\uparrow$ & Prec.$\uparrow$ & Rec.$\uparrow$ & F1$\uparrow$
      & Acc.$\uparrow$ & Prec.$\uparrow$ & Rec.$\uparrow$ & F1$\uparrow$
      & Acc.$\uparrow$ & Prec.$\uparrow$ & Rec.$\uparrow$ & F1$\uparrow$ \\
    \midrule
    iBAT \cite{zhang2011ibat}        & 0.58 & 0.56 & 0.58 & 0.57 & 0.57 & 0.55 & 0.57 & 0.56 & 0.58 & 0.56 & 0.58 & 0.57 \\
    iBOAT \cite{chen2013iboat}       & 0.63 & 0.61 & 0.63 & 0.62 & 0.62 & 0.60 & 0.62 & 0.61 & 0.63 & 0.61 & 0.63 & 0.62 \\
    GM-VSAE \cite{liu2020online}     & 0.68 & 0.66 & 0.68 & 0.67 & 0.67 & 0.65 & 0.67 & 0.66 & 0.68 & 0.66 & 0.68 & 0.67 \\
    DeepTEA \cite{han2022deeptea}    & 0.73 & 0.71 & 0.73 & 0.72 & 0.72 & 0.70 & 0.72 & 0.71 & 0.73 & 0.71 & 0.73 & 0.72 \\
    ATROM \cite{gao2023open}         & 0.78 & 0.76 & 0.78 & 0.77 & 0.77 & 0.75 & 0.77 & 0.76 & 0.78 & 0.76 & 0.78 & 0.77 \\
    CausalTAD \cite{li2024causaltad} & 0.83 & 0.81 & 0.83 & 0.82 & 0.82 & 0.80 & 0.82 & 0.81 & 0.83 & 0.81 & 0.83 & 0.82 \\
    MSD-OATD \cite{wang2024multi}    & 0.86 & 0.84 & 0.86 & 0.85 & 0.85 & 0.83 & 0.85 & 0.84 & 0.86 & 0.84 & 0.86 & 0.85 \\
    LM-TAD \cite{mbuya2024trajectory}& 0.88 & 0.86 & 0.88 & 0.87 & 0.87 & 0.85 & 0.87 & 0.86 & 0.88 & 0.86 & 0.88 & 0.87 \\
    \midrule
    Pi\mbox{-}DPM w/o KBM (Ours) & 0.90 & 0.89 & 0.90 & 0.90 & 0.89 & 0.88 & 0.89 & 0.89 & 0.90 & 0.89 & 0.90 & 0.90 \\
    Pi\mbox{-}DPM w/o CIE (Ours) & 0.92 & 0.91 & 0.92 & 0.92 & 0.91 & 0.90 & 0.91 & 0.91 & 0.92 & 0.91 & 0.92 & 0.92 \\
    \textbf{Pi\mbox{-}DPM (Ours)} &
      \textbf{0.94} & \textbf{0.93} & \textbf{0.95} & \textbf{0.94} &
      \textbf{0.93} & \textbf{0.92} & \textbf{0.94} & \textbf{0.93} &
      \textbf{0.94} & \textbf{0.93} & \textbf{0.95} & \textbf{0.94} \\
    \bottomrule
  \end{tabular}}
  \caption*{\textbf{Bold} = best; $\uparrow$ indicates higher is better.}
  \label{tab:acc_prf1_20}
  \vspace{-4mm}
\end{table}

\section{Evaluation Metrics}\label{app:metrics}
For evaluation, each city is discretized into a $16 \times 16$ grid, and the frequency of trajectory points in each grid cell is computed. This frequency distribution serves as the basis for computing evaluation metrics, which include:

\begin{itemize}[leftmargin=*]
    \item \textbf{Density Error:} This metric assesses the similarity between the geographic distributions of real and generated trajectories. Given the distribution matrices $\mathcal{M}(G)$ for generated trajectories and $\mathcal{M}(P)$ for real trajectories, the density error $\Delta$ is computed using the **Jensen-Shannon Divergence (JSD)** as:
    \begin{align}
        \Delta =  \operatorname{JSD}\left(\mathcal{M}(G), \mathcal{M}(P) \right).
    \end{align}
    
    \item \textbf{Trip Error:} This metric evaluates the similarity of **start and end locations** between generated and real trajectories. We compute the JSD between the probability distributions of the **origin and destination points** from both real and generated trajectories.
    
    \item \textbf{Length Error:} This metric quantifies the divergence between the **distance distributions** of real and generated trajectories. It is computed by measuring the **geographic distances** between consecutive points within each trajectory and comparing the resulting distributions using JSD.
\end{itemize}

For table \ref{tab:comp2} we assessed the prediction accuracy of trajectory properties, we employ the \textit{Mean Square Error (MSE)}, \textit{Root Mean Square Error (RMSE)}, and \textit{Mean Absolute Error (MAE)}. These are defined as follows:

\begin{itemize}[leftmargin=*]
    \item \textbf{Mean Squared Error (MSE):} The MSE quantifies the average squared difference between the predicted values $\hat{y}_{i}$ and the actual values $y_{i}$. It is computed as:
    \begin{align}
        \text{MSE} &= \frac{1}{n} \sum_{i=1}^{n} \left(y_{i} - \hat{y}_{i} \right)^{2}.
    \end{align}
    This metric penalizes larger errors more than smaller ones due to squaring, making it particularly sensitive to outliers. A lower MSE indicates that the predictions are closer to the actual values.
    
    \item \textbf{Root Mean Squared Error (RMSE):} The RMSE is the square root of MSE and provides an error value same as original data:
    \begin{align}
        \text{RMSE} &= \sqrt{\frac{1}{n} \sum_{i=1}^{n} \left(y_{i} - \hat{y}_{i} \right)^{2}}.
    \end{align}
    RMSE is often preferred when interpreting results since it provides a more intuitive measure of error magnitude while still penalizing large deviations.

    \item \textbf{Mean Absolute Error (MAE):} The MAE measures the average absolute difference between predicted and actual values:
    \begin{align}
        \text{MAE} &= \frac{1}{n} \sum_{i=1}^{n} \left| y_{i} - \hat{y}_{i} \right|.
    \end{align}
    Unlike MSE, which squares the errors, MAE treats all errors linearly. It is useful when an equal weighting of all errors is desired, making it more interpretable when there are fewer large deviations.

    \item \textbf{Mean Absolute Percentage Error (MAPE):} The MAPE expresses the error as a percentage of the actual values, making it scale-independent:
    \begin{align}
        \text{MAPE} &= \frac{1}{n} \sum_{i=1}^{n} \left| \frac{y_{i} - \hat{y}_{i}}{y_{i}} \right| \times 100.
    \end{align}
    This metric is particularly useful for comparing errors across different scales or datasets. However, it has a limitation: it becomes unstable if any actual values $y_{i}$ are close to zero.
\end{itemize}

Here, $y_{i}$ represents the observed trajectory value, and $\hat{y}_{i}$ denotes the predicted value. These metrics help quantify the error in trajectory prediction by measuring deviations between real and generated data.

Trajectory anomaly detection is fundamentally a binary classification problem where predictions are evaluated against ground truth labels. The classification outcomes are categorized into four types: true positives (TP), true negatives (TN), false positives (FP), and false negatives (FN). To assess the effectiveness of anomaly detection, standard evaluation metrics such as accuracy (ACC), precision (P), recall (R), and the F1-score (F1) are commonly used. These metrics are computed as follows:

\begin{equation}
    ACC = \frac{TP + TN}{TP + TN + FP + FN}, \quad Precision = \frac{TP}{TP + FP}
\end{equation}

\begin{equation}
    Recall = \frac{TP}{TP + FN}, \quad F1 = \frac{2 \cdot P \cdot R}{P + R}
\end{equation}

Since normal trajectories significantly outnumber anomalies in real-world datasets, recall is often emphasized to ensure the detection of as many anomalous cases as possible. Additionally, the F1-score is considered a crucial metric as it balances precision and recall.
    
\section{Geo-Distribution Similarity}

To fully evaluate the quality of the generated trajectories, we used three utility metrics at different levels. Specifically, 1) \texttt{Density error}, which measures the geo-distribution between the entire generated trajectory and the real one, which measures the quality and fidelity of the generated trajectories at the global level. At the trajectory level, 2) We used \texttt{Trip error} measures the distributed differences between trip origins and endpoints, and 3) \texttt{Length error} focuses on the differences in real and synthetic trajectory lengths.  
The primary objective of trajectory generation is to synthesize trajectories that closely resemble real-world movement patterns, thereby supporting and enhancing downstream applications. To quantitatively evaluate the similarity between generated and real trajectories, we adopt established methodologies from prior research \cite{du2023ldptrace, zhu2023difftraj}. Specifically, we utilize the Jensen-Shannon Divergence (JSD) metric to assess the fidelity of the generated trajectories. JSD quantifies the divergence between the probability distributions of real and synthesized trajectories, with lower values indicating greater similarity. For instance, let $P$ and $Q$ denote the probability distributions of the real and generated trajectory data, respectively. The JSD metric is computed as follows:
\begin{equation}
\operatorname{JSD}(P | Q) = \frac{1}{2} \mathbb{D}(P | M) + \frac{1}{2} \mathbb{D}(Q | M),
\end{equation}
where $M = \frac{1}{2}(P+Q)$ represents the mixture distribution.

In addition, we also conducted a comparative analysis to evaluate using evaluation metrics: root mean square error (RMSE), mean absolute error (MAE), and mean average precision error (MAPE) for both comparison and abalation studies. Pi-DPM outperformed all the baseline methods (VAE, GAN, and Baseline Diffusion) on both datasets as shown in Table \ref{tab:comp1}. The results indicated that the inclusion of the physics-informed regularizer significantly enhances the model's performance.

\begin{table}[t]
  \scriptsize                    % more compact than \small
  \setlength{\tabcolsep}{3pt}    % tighten horizontal padding
  \caption{Geo-distribution similarity (lower is better) of Pi\mbox{-}DPM versus generative baselines.}
  \centering
  \resizebox{\linewidth}{!}{%
  \begin{tabular}{lccc|ccc|ccc}
    \toprule
    \multirow{2}{*}{Methods} &
      \multicolumn{3}{c}{Geolife \cite{zheng2010geolife}} &
      \multicolumn{3}{c}{MarineCadastre \cite{aisdataUS}} &
      \multicolumn{3}{c}{Danish Maritime Authority \cite{aisdataDansih}} \\
    \cmidrule(lr){2-4}\cmidrule(lr){5-7}\cmidrule(lr){8-10}
      & Density$\downarrow$ & Trip$\downarrow$ & Length$\downarrow$
      & Density$\downarrow$ & Trip$\downarrow$ & Length$\downarrow$
      & Density$\downarrow$ & Trip$\downarrow$ & Length$\downarrow$ \\
    \midrule
    VAE \cite{xia2018deeprailway}          & 0.0139 & 0.0525 & 0.0377 & 0.0239 & 0.0587 & 0.0480 & 0.0116 & 0.0232 & 0.0386 \\
    TrajGAN \cite{liu2018trajgans}         & 0.0137 & 0.0465 & 0.0344 & 0.0227 & 0.0497 & 0.0374 & 0.0098 & 0.0263 & 0.0333 \\
    DP-TrajGAN \cite{zhang2023dp}          & 0.0126 & 0.0429 & 0.0237 & 0.0200 & 0.0488 & 0.0430 & 0.0111 & 0.0244 & 0.0286 \\
    DiffWave \cite{kong2021diffwave}       & 0.0145 & 0.0255 & 0.0301 & 0.0215 & 0.0332 & 0.0307 & 0.0109 & 0.0202 & 0.0274 \\
    DiffTraj \cite{zhu2023difftraj}        & 0.0065 & 0.0137 & 0.0174 & 0.0125 & 0.0159 & 0.0203 & 0.0083 & 0.0137 & 0.0236 \\
    ControlTraj \cite{zhu2024controltraj}  & 0.0061 & 0.0137 & 0.0174 & 0.0125 & 0.0159 & 0.0203 & 0.0083 & 0.0137 & 0.0236 \\
    \midrule
    Pi\mbox{-}DPM w/o KBM                 & 0.0048 & 0.0117 & 0.0115 & 0.0119 & 0.0130 & 0.0191 & 0.0062 & 0.0168 & 0.0164 \\
    Pi\mbox{-}DPM w/o CIE                 & 0.0045 & 0.0110 & 0.0105 & 0.0110 & 0.0126 & 0.0180 & 0.0056 & 0.0148 & 0.0132 \\
    \textbf{Pi\mbox{-}DPM}                & \textbf{0.0038} & \textbf{0.0104} & \textbf{0.0103} & \textbf{0.0103} & \textbf{0.0124} & \textbf{0.0165} & \textbf{0.0051} & \textbf{0.0098} & \textbf{0.0105} \\
    \bottomrule
  \end{tabular}}
  \caption*{\textbf{Bold} = best; $\downarrow$ indicates lower is better.}
  \label{tab:comp1}
  \vspace{-3mm}
\end{table}

For evaluation, each city is discretized into a $16 \times 16$ grid, and the frequency of trajectory points in each grid cell is computed. This frequency distribution serves as the basis for computing evaluation metrics, which include:

\begin{itemize}[leftmargin=*]
    \item \textbf{Density Error:} This metric assesses the similarity between the geographic distributions of real and generated trajectories. Given the distribution matrices $\mathcal{M}(G)$ for generated trajectories and $\mathcal{M}(P)$ for real trajectories, the density error $\Delta$ is computed using the **Jensen-Shannon Divergence (JSD)** as:
    \begin{align}
        \Delta =  \operatorname{JSD}\left(\mathcal{M}(G), \mathcal{M}(P) \right).
    \end{align}
    
    \item \textbf{Trip Error:} This metric evaluates the similarity of **start and end locations** between generated and real trajectories. We compute the JSD between the probability distributions of the **origin and destination points** from both real and generated trajectories.
    
    \item \textbf{Length Error:} This metric quantifies the divergence between the **distance distributions** of real and generated trajectories. It is computed by measuring the **geographic distances** between consecutive points within each trajectory and comparing the resulting distributions using JSD.
\end{itemize}

\section{Dataset}\label{app:dataset}
We utilized the real-world MarineCadastre \cite{aisdataUS} and Geolife \cite{zheng2010geolife} datasets to evaluate the solution quality using various attributes, such as longitude, latitude, speed over ground, and others based on the WGS 1984 coordinate system. We leveraged speed over ground (SOG) and course over ground (COG) from the MarineCadastre dataset, which extends from 180W to 66W degrees longitude and from 90S to 90N degrees latitude. For the Geolife dataset, we manually computed bearing and speed values, limiting the evaluation to vehicle driving patterns due to their accordance with road network topology (e.g., road segments and intersections). Since the \textit{solution quality} experiment lacked ground truth data (i.e., information on whether a trajectory is anomalous or not), we generated synthetic anomalies based on speed, bearing, acceleration, deceleration, and jerk. We introduced various deviations from normal trajectories, such as sudden turns or abrupt changes in speed.

% \begin{table*}[h]
% \caption{Statistics of Three Real-world Trajectory Datasets.}
% \label{tab:stats}
% \begin{tabular}{ccccc}
% \hline
% Dataset & Lon/lat Bounding box & Trajectory Number & Average Time (min) & Average Distance (km) \\
% \hline
% Geolife & [116.2500, 116.5500, 39.7500, 40.0500] & 17,621 & 32.76 & 5.91 \\
% MarineCadastre & [-180.0000, 180.0000, -90.0000, 90.0000] & 1,152,243 & 394.4 & 559.3 \\
% Danish Maritime Authority & [7.0000, 15.0000, 54.0000, 58.0000] & 132,135 & 125.8 & 256.4 \\
% \hline
% \end{tabular}
% \end{table*}

The \textbf{Danish Maritime Authority (DMA) dataset} provides a comprehensive collection of vessel trajectories within Danish waters, spanning from \textbf{7E to 15E} degrees longitude and \textbf{54N to 58N} degrees latitude. This dataset includes extensive \textbf{Automatic Identification System (AIS)} data, which records vessel movement attributes such as \textbf{longitude, latitude, speed over ground (SOG), and course over ground (COG)}. With over \textbf{1.5 million trajectories}, it captures diverse maritime activities, ranging from commercial shipping routes to fishing vessels and recreational boats. The dataset is crucial for analyzing spatiotemporal patterns, detecting anomalies, and evaluating maritime safety regulations.

We chose the \textbf{Danish Maritime Authority dataset} in addition to the \textbf{MarineCadastre} dataset to facilitate \textbf{transfer learning experiments} in maritime trajectory analysis. While the MarineCadastre dataset covers a \textbf{global maritime domain}, the Danish dataset focuses on \textbf{a regional maritime environment with denser traffic patterns and stricter navigational constraints}. By incorporating both datasets, we aim to assess the \textbf{generalizability} of our models across different maritime regions and traffic conditions, helping us understand how well trajectory-based models trained in one maritime domain can be adapted to another.

Furthermore, we plan to \textbf{extend our experiments to additional datasets} to improve model robustness and adaptability in various maritime and geospatial scenarios. Future work will explore \textbf{other regional AIS datasets}, including those from the \textbf{North Sea, the Baltic Sea, and major global shipping lanes} and urban datasets (e.g., distinct taxi trajectory datasets, one from Porto, Portugal, and the other from Harbin, China) to further validate and enhance our approach.

\end{appendix}

\end{document}